\documentclass[journal=jacsat,manuscript=article]{achemso}

\usepackage[version=3]{mhchem} 
\usepackage{amsmath}
\usepackage{amssymb}
\usepackage{adjustbox}
\usepackage{graphicx}
\usepackage{subcaption}
\usepackage{fancybox}  
\newcommand{\desc}[1]{{\small``#1''}}
\usepackage{subcaption}
\usepackage{fancybox}  

\usepackage{longtable}
\usepackage{tabularx}
\usepackage{geometry}
\author{Mrityunjay Sharma}
\affiliation[CSIR]
{CSIR- Central Scientific Instruments Organisation, Sector 30-C, Chandigarh-160030, India}
\alsoaffiliation[ACSIR]
{Academy of Scientific and Innovative Research (AcSIR), Ghaziabad-201002, India}
\alsoaffiliation[Prof]
{Department of Higher Education,Himachal Pradesh,Shimla -171001,India}
\author{Sarabeshwar Balaji}
\affiliation[IISERB]
{Indian Institute of Science Education and Research Bhopal(IISERB),
Madhya Pradesh-462066,India}
\author{Pinaki Saha}
\affiliation[UHBIO]
{University of Hertfordshire, UH Biocomputation Group,United Kingdom}
\author{Ritesh Kumar}
\email{riteshkr@csio.res.in}
\affiliation[CSIR]
{CSIR- Central Scientific Instruments Organisation, Sector 30-C, Chandigarh-160030, India}
\alsoaffiliation[ACSIR]
{Academy of Scientific and Innovative Research (AcSIR), Ghaziabad-201002, India}

\title[An \textsf{achemso} demo]
  {Navigating the Fragrance space Via Graph Generative Models And Predicting Odors}

\abbreviations{IR,NMR,UV}
\keywords{American Chemical Society, \LaTeX}

\begin{document}

\begin{abstract}
We explore a suite of generative modelling techniques to efficiently navigate and explore the complex landscapes of odor and the broader chemical space. Unlike traditional approaches, we not only generate molecules but also predict the odor likeliness  with ROC AUC score of
0.97 and assign probable odor labels. We correlate odor likeliness with physicochemical features of molecules using machine learning techniques 
and leverage SHAP (SHapley Additive exPlanations) to demonstrate the interpretability of the function. The whole process involves four key stages: molecule generation, stringent sanitization checks for molecular validity, fragrance likeliness screening and odor prediction of the generated molecules. By making our code and trained models publicly accessible, we aim to facilitate broader adoption of our research across applications in fragrance discovery and olfactory research.
\end{abstract}

\section{Introduction}
The generation of novel molecules is a cornerstone of chemical engineering, essential for developing compounds that meet specific design goals. However, this process is often labor intensive, time-consuming and primarily based on trial and error.
Recent advances in AI, machine learning, and deep learning, particularly in graph neural networks (GNNs)\cite{scarselli2008graph}, have paved the way for developing chemically stable and valid molecules by efficiently analyzing and embedding atomic as well as molecular features\cite{reiser2022graph}. 
GNNs use message passing to study neighboring nodes through homophily, which reflects the tendency of nodes to align with similar neighbors and influence, which highlights how connections between atoms affect their individual properties.
This capability enables a more efficient exploration of the expansive virtual chemical space, surpassing traditional virtual screening approaches and additionally also offers a finer navigation\cite{alves2022graph}. 
While generative models and GNN-based approaches have revolutionized areas such as  biomarker development\cite{desaire2022not}, chemical simulation \cite{meuwly2021machine}, spectroscopy and analytical method development \cite{dos2023unraveling}, digital material design \cite{chan2022application} and drug discovery\cite{david2020molecular,saifi2024artificial},   their potential remains untapped in olfaction, where the complex structure-odor relationship poses unique challenges.

Molecules with different structures can exhibit the same odor, while molecules with similar structures can produce distinct odors, making the relationship between structure and odor difficult to decipher\cite{sell2006unpredictability}. This ongoing challenge has been explored using various approaches such as rough set theory, graph neural network and deep learning \cite{radhakrishnapany2020design,rodrigues2024harnessing,lee2023principal}.  Inspired by drug discovery where we study the quantitative estimate of drug likeness QED\cite{bickerton2012quantifying}, researchers have linked fragrance likelines with molecule’s boiling point, vapour pressure, lipophilicity, and volatility\cite{mayhew2022transport}. Moreover there has been notable research on understanding the physicochemical features of fragrant molecules, offering insights that can complement generative and AI-driven approaches, \cite{mayhew2022transport,lepoittevin1998molecular,ruddigkeit2012enumeration,ruddigkeit2014expanding}. Integrating these approaches can expand labeled odor molecule databases, generate novel compounds, reduce experimentation, and lower costs\cite{rodrigues2024molecule}.

Building on the insights from previous approaches, we propose a more practical approach to fragrant molecule design, which involves constraining the process to focus on physicochemical and molecular properties rather than just molecule structures, which most earlier approaches did. In our study, models are trained on a curated dataset\cite{aryan_amit_barsainyan_ritesh_kumar_pinaki_saha_michael_schmuker_2023} of odorous molecules to demonstrate the utility and versatility of various graph generative models. This study aggregates key attributes into a unified framework and proposes an  odor-likeliness equation, and assigns odor labels to probable odorous molecules based on the equation. We validate the performance of the generative models according to the MOSES benchmarks\cite{polykovskiy2020molecular}.
\begin{figure}[htbp]
    \centering
    \includegraphics[width=1.0\linewidth]{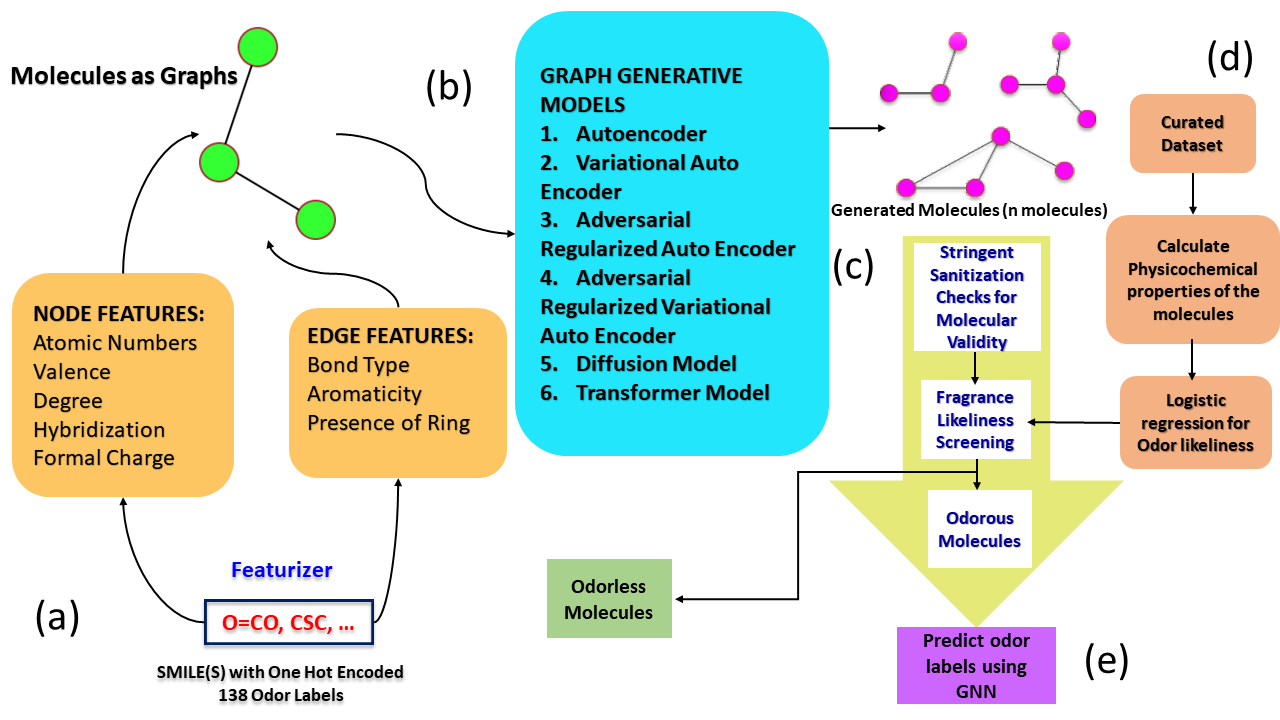} 
    \caption{\textbf{Overview of the methodology}: \textbf{(a)} From odorous molecules of the curated dataset \cite{aryan_amit_barsainyan_ritesh_kumar_pinaki_saha_michael_schmuker_2023}  the edge and node features are extracted. The SMILE strings are then converted into a Pytorch geomteric dataset. \textbf{(b)} Generative models are applied thereafter and graphs are generated. \textbf{(c)} Generated graphs are checked for molecular stability based on the node and edge features. Molecules are constructed from the graph obtained  and then converted into the SMILE strings. Again they are checked in the PubChem database\cite{kim2023pubchem}. \textbf{(d)} The valid and novel SMILE strings are checked for odor likeness using equation which is built using physicochemical features. \textbf{(e)} In case generated molecules are valid and odorous, the odor is predicted by graph neural network based model.}
    \label{fig:methodology}
\end{figure}
\subsection{Graph Generative Models}
In this research, we have generated molecules using six different generative models,  namely Graph AutoEncoder (GAE),
Variational Graph AutoEncoder (VGAE), Transformer model, Diffusion model, Adversarially Regularized Graph Autoencoder (ARGA) and
Adversarially Regularized Graph Variational Autoencoder (ARGVA). Throughout the explanation in this section we will consider a graph G = (V, E) with  $\lvert V \rvert =$ n nodes and  $\lvert E \rvert =$ m
edges.
\subsubsection{Graph AutoEncoder (GAE)}
GAE refers to a type of neural network that operates in an unsupervised manner. It consists of two main components: an encoder and a decoder. The encoder maps or embeds an input X, where $v_i \in V$, and the input has n dimensions, into a lower-dimensional latent space $z_i \in {R}^d, \text{ with d} \ll n$. The decoder reconstructs the input from this latent representation \cite{kipf2016variational}. This technique of representing the input in lower dimension can also be used for compression and some other ML algorithm can also use this input which has a lower dimension. Although Principal Component Analysis (PCA) and encoders may appear similar, they vary in that PCA only captures linear relationships, whereas encoders can model complex, non-linear relationships. Training a GAE basically involves reducing the reconstruction loss by backpropagation. GAE suffers from limitations, such as its latent space is not continuous and lacks symmetry around the origin\cite{kipf2016variational}.

\subsubsection{Variational Graph AutoEncoder (VGAE)} 
Unlike  GAE which maps the input X to a point in the embedding space, VGAE maps the input to a multivariate normal distribution. The encoder maps each input to a mean vector($\vec{\mu}$) and variance vector($\vec{\sigma^2}$). To avoid using an extra activation function, $\log \vec{\sigma^2}$,which can also be negative is used instead of ($\vec{\sigma^2}$), which is always positive. After obtaining \(\mu\) and ($\vec{\sigma^2}$), we aim to make both \(\mu\) and  \(\sigma\) approach to 1 or \(\log \sigma^2\) approach 0 which makes the final distribution approximate \(N(0,1)\). To generate the embedding \(z\), we compute \(z = \mu + \sigma \cdot \epsilon\), where \(\epsilon \sim N(0,1)\). This process is known as the reparameterization trick \cite{kingma2013auto}. Finally, with the latent variable \(z\), the output \(\hat{X}\) is generated using the decoder\cite{kipf2016variational}. VGAE uses reconstruction error along with Kullback-Leibler(KL) divergence. KL divergence quantifies the difference between normal distribution generated from standard normal distribution. This is a strategy by which we can regularize the latent space and enforce it to match a distribution.

\subsubsection{Adversarially Regularized Graph Autoencoder (ARGA)}

Unlike the previous models ARGA works in an adversarial manner, the main components are generator and discriminator\cite{goodfellow2014generative}. The generator generates fake data and the discriminator distinguishes whether latent code is from an arbitrary prior distribution or from the encoder. In conjunction to the reconstruction loss there is adversarial loss from the discriminator. These models tend to regularize the latent space and improve the realism of the generated samples. If A is the original graph, X is the node content and $\hat{A}$ is the reconstructed graph. 
Training is performed using the following equation for encoder \cite{pan2018adversarially}:-

\begin{equation}
\min_G \max_D \mathbb{E}_{z \sim p_z} \big[\log \mathcal{D}(Z)\big] + \mathbb{E}_{x \sim p(x)} \big[\log\big(1 - \mathcal{D}(G(\mathbf{X}, \mathbf{A}))\big)\big].
\end{equation}
where $\mathcal{D}(Z)$ and  
 G($\mathbf{X}, \mathbf{A}$) indicate the discriminator and generator explained above. $\mathcal{D}(Z)$ is in [0,1]
It can be observed that the first term is large when $\mathcal{D}(Z)]$ is close to 1, high value to output generated which look real. The second term is large if $\mathcal{D}(Z)]$ is close to zero, in other words $1-\mathcal{D}(Z)]$ is large, assigning low value to fake outputs generated. For a graph data the reconstruction error is minimized as follows :-
\begin{equation}
\mathcal{L}_0 = \mathbb{E}_{q(\mathbf{Z}|\mathbf{X},\mathbf{A})}[\log p(\mathbf{A}|\mathbf{Z})]
\end{equation}
\subsubsection{Adversarially Regularized Graph Variational Autoencoder (ARVGA)}

The ARGVA is very similar to ARGA, but the key difference is that while ARGA uses a graph convolutional autoencoder, ARGVA employs a variational graph autoencoder. In ARGA, there is only a single embedding, whereas in ARGVA, we generate both a mean and standard deviation for the embeddings. Similar to how the loss in a Graph Autoencoder (GAE) is purely a reconstruction loss and in a Variational Graph Autoencoder (VGAE) the loss includes both reconstruction and KL regularization, the same distinction applies here. Real samples are drawn from a Gaussian distribution, while fake samples come from the latent encodings. The setup mirrors that of a VAE, with an adversarial component attached \cite{pan2018adversarially}.

\subsubsection{Diffusion model}

A diffusion probabilistic model, sometimes known as a "diffusion model," is a parameterised Markov chain trained with variational inference to generate data-matching samples over a finite time period. Since the diffusion model has given promising results in generating high quality images\cite{ho2020denoising} we also experimented with the potential of a diffusion model for graph generation. It generates the data by progressively denoising the noise. Starting from pure Gaussian noise, it iteratively denoise the sample using a trained network. The network predicts and removes the noise step by step and eventually produces a clean graph. The formula of denoising is based on reverse diffusion .
\begin{equation}
x_{t-1} = \frac{1}{A_t} \left(x_t - \frac{1-\alpha_t}{\sqrt{1-\alpha_t}} e_0 (x_t, t)\right) + \sigma_z z
\end{equation}
Where, $x_t$ is the noisy image at step $t$, $\alpha_t$ and $\sqrt{1-\alpha_t}$ are parameters derived from the noise schedule, $\epsilon_\theta(x_t, t)$ is the predicted noise by the neural network, $\sigma_t$ is the standard deviation for the noise added during each reverse diffusion step, $z$ is the Gaussian noise. This formula denoisifies the image $x_t$ to recover the clean image $x_0$.

\subsubsection{Transformer model}

Message passing in GNNs typically focuses on feature propagation without considering unlabelled instances. The Unified Message Passing Model (UniMP) incorporates both labels and node features in both training and prediction stages\cite{shi2020masked}. It transforms partial node labels from one-hot to dense vectors using embeddings. A hyper paramter label\_rate controls the masked label prediction applied to optimize this process, , similar to BERT, where input words are masked and used for prediction \cite{devlin2018bert}.  A multi-layer Graph Transformer network propagates information between nodes, enabling each node to aggregate both feature and label information from its neighbours. It's also observed that label propagation (LPA) combined with GNN improves efficiency. While transformers have been widely used in NLP, only vanilla multi-head attention has been adapted for graph learning, taking into account edge features. The unique aspect of UniMP is its use of label propagation, implemented via transformers.

All the models presented in this paper employ deep learning embeddings, built by stacking layers of neural networks of varied dimensions. Graph Autoencoder (GAE), focus on preserving similarity and minimizing reconstruction error but do not regularize the embedded space. We also implemented ARGA and ARGVA, which have a regularized latent space and constrained to follow a specific prior distribution. Diffusion and transformer models often operate in higher-dimensional spaces and capture more complex interactions. Transformer models excel in capturing long-range dependencies leveraging attention mechanisms which assists in separating features.
VAE regulates regularization through KL divergence, ARGA relies on adversarial training, ARGVA combines both adversarial and variational methods, diffusion models achieve regularization by adding noise and transformer models use attention mechanisms.
\subsection{Logistic Regression}
A popular supervised machine learning model for classification problems is logistic regression. It employs a sigmoid function to map independent variables on a scale between 0 and 1. Linear regression fits a straight line and is used for predicting quantitative outcomes. In contrast logistic regression fits a sigmoid curve, is suitable for predicting probabilities and is used for qualitative or categorical outcomes\cite{james2013introduction}. Its three types:  binary, multinomial, and ordinal are based on the nature of the target variable:

In this work, we use binary logistic regression to check the odor likelihood. Logistic regression is mathematically represented as:

\[
\log \frac{p(X)}{1 - p(X)} = \beta_0 + \beta_1 X
\]

Here, \(\frac{p(X)}{1 - p(X)}\) represents the odds, \(\log \frac{p(X)}{1 - p(X)}\) is the log-odds (logit), and \(\beta_0, \beta_1\) are model coefficients. The odds range from 0 to \(\infty\), where values close to 0 indicate low probability, and values close to \(\infty\) indicate high probability. 

Logistic regression has been extensively applied recently in various scientific fields, including analyzing molecular dynamics simulations\cite{wu2024allosteric}, estimating amino acid contributions\cite{sugiki2022logistic}, studying substitution reaction mechanisms\cite{dood2020analyzing} and many more. In this work, it is utilized to predict odor likeliness by classifying molecules as either odorous or odorless (\(y \in \{0, 1\}\)).

\section{Methods}

\subsection{Dataset creation}
We used a curated dataset combining the GoodScents\cite{luebke2019good} and Leffingwell PMP 2001\cite{leffing} datasets, both of which contain odorant molecules and their associated odor descriptors \cite{aryan_amit_barsainyan_ritesh_kumar_pinaki_saha_michael_schmuker_2023,lee2023principal}. This dataset has 4,983 molecules all of which have multilabeled odor labels associated and also includes odorless molecules. For this work, we used 4,751 chemically stable and odorous molecules out of the 4,983 available, which were subsequently employed in the generative models. The SMILES strings in the curated dataset were converted into a PyTorch Geometric dataset, incorporating node and edge attributes. Feauturizer was applied where the node feature length is 134 for the corresponding SMILES in the dataset, which were represented as graphs. This length is achieved by concatenating the following features (one hot encoded): valence (0-6), degree (0-5), number of hydrogens (0-4),formal charge (-1 to 2), atomic number (first 100 atoms), hybridization (sp, sp2, sp3, sp3d, sp3d2) and one extra bit for other values not defined. The edge features are 6-bit, comprising of: single, double, triple, or aromatic, is ring and one extra bit for any other condition\cite{lee2023principal}.
 Graph generative models were then applied to generate graphs, which were subsequently converted into molecules. 
\subsection{Benchmarks}
The generated molecules were screened for instability, invalid valency, chemical stability, unrealistic state and too many hydrogen using a custom filtering mechanism alongside  the sanitize function in the RDKit to ensure validity. In order to ascertain that each molecule has a unique representation and to address graph isomorphism issue, the process of SMILES canonicalization was applied. Model evaluation was done by employing metrics and calculation methods as defined in the MOSES benchmark \cite{polykovskiy2020molecular}. 
These include validity (adherence to chemical rules), uniqueness (absence of duplication among the generated SMILES), novelty (proportion of generated molecules absent from the training set), scaffold similarity (Scaff), which is akin to the fragment similarity metric but compares frequencies of Bemis–Murcko scaffolds \cite{bemis1996properties}, and similarity to a nearest neighbor (SNN). The limits of Scaff
are [0,1], where low values indicate the rare production of a certain chemotype from the training set.  The limits of SNN
are [0,1], where low values indicate that the generated molecules are far from the manifold of the reference set, resulting in low similarity to the nearest neighbor.  Diversity was also calculated using pairwise Tanimoto similarities\cite{bajusz2015tanimoto}, with a score of $\textit{1 - \text{average\_similarity}}$, ranging from 0 (identical) to 1 (dissimilar), ensuring the generation of diverse and unique molecules. We validated the generated molecules by cross-referencing them with PubChem\cite{kim2023pubchem} to ensure chemical stability and practical usability. Only molecules found in PubChem were flagged as valid. Our models were evaluated based on the benchmarks and the variety of odor labels generated.

\subsection{Odor Likeliness criteria}
There cannot be a single parameter, such as volatility, that can predict the odor likeliness of a molecule. While volatility allows a molecule to transmit scent to olfactory receptors, it must also be hydrophilic enough to absorb into the mucus in the olfactory epithelium and bind to specific olfactory receptors. Previous studies have navigated fragrance space using boiling points and vapor pressures, which are difficult to obtain, and proposed rules such as the “rule of three”\cite{mayhew2022transport} which states that molecules with molecular weights between 30 and 300 Da and fewer than three heteroatoms tend to be odorous. Evidence also supports a ``fragrance-like'' (FL) property range defined as molecules with heavy atom count(HAC) $\leq21$, composed of C, H, O, or S, with a maximum of three hetero atoms (S + O $\leq$ 3) and one hydrogen-bond donor (HBD $\leq$ 1) for molecules to be odorous  \cite{ruddigkeit2014expanding} . Researchers have also classified molecules as odorous and odorless based on HAC $\leq$ 17, restricting the atoms to only organic molecules and elements such as C, H, O, N, S, and halogens to build a database which is called GDB-17 \cite{ruddigkeit2014expanding}. In order to evaluate the odor likeliness criteria, we compare the results obtained from the proposed method with those explained above.

In this study, logistic regression model is applied to curated fragrance dataset \cite{aryan_amit_barsainyan_ritesh_kumar_pinaki_saha_michael_schmuker_2023}. This model predicts whether a molecule has a odor based on log-odds transforming probabilities into a linear combination of features. The properties of a novel molecule are not known beforehand, so they must be calculated empirically. We used physicochemical and molecular properties such as molecular weight, LogP( logarithm of the partition coefficient between octanol and water), Rotatable Bonds, FCFP4 count(part of the Fingerprinting Convolutional Family (FCFP)), fraction of Sp2 Hybridized Atoms etc. Table S1 in the supplementary section lists all the features used to build the odor likliness critera.  All these features were calculated through empirical methods using built-in functions in the RDKit library in Python. To address class imbalance, we used the Synthetic Minority Over-sampling Technique (SMOTE) \cite{maklin2022synthetic}. Multicollinearity was reduced by removing highly correlated features (correlation $>$ 0.75) and further variables having Variance Inflation Factor (VIF) $>$ 5 \cite{chan2022mitigating} were eliminated, leaving 20 attributes.  After feature selection, the top five most important features were identified using SHAP (SHapley Additive exPlanations)\cite{NIPS2017_7062} values, based on their importance and a logistic regression equation was built.
\subsection{Hyperparameter tuning}
To tune the hyperparameters for all the models we utilized Optuna \cite{optuna_2019}, a Bayesian optimization framework. Table \ref{tab:hyperparameters} presents a detailed view of the search values and chosen values of hyperparameter values for all the models. We conducted a total of 200 such hyperparameter trial per model with 60 epcohs. Early stopping was implemented to reduce excessive training cycles by terminating the training loop when the validation loss failed to improve within a specified patience interval. Patience value represented the number of epochs to wait for validation loss improvement and also ensured balance between convergence time and computational efficiency. For adversarial models such as ARGA, early stopping also monitored adversarial loss stability, thereby ensuring equilibrium between the discriminator and generator. Early stopping approach helped to prevent overfitting and also facilitated faster convergence significantly enhancing the efficiency of hyperparameter tuning. The hyperparameters analysed in this research included hidden dimensions, batch size, latent dimensions, learning rate and patience. 
\begin{table}[htbp]
\centering
\resizebox{\textwidth}{!}{%
\begin{tabular}{|l|c|c|c|c|c|c|c|}
\hline
 & Search Space & GAE & VGAE & ARGA & ARGVA &  Diffusion model & Transformer \\
\hline
Hidden Dim     &  \{64, 512\} & 320  & 256 & 192  &384  &   448 & 320 \\
\hline
Latent Dim     & \{32, 128\}   & 128  & 32 &  64 & 128  &  128 & 96 \\
\hline
Learning Rate  & \{ 1e-5, 1e-1\}  &  1.3e-3 & 4.3e-3& 1.3e-3 &1.8e-2  & 
   8.6e-4 & 1.7e-4
 \\
\hline
Batch Size     &\{ 8, 16, 32, 64\}   & 16   & 32   &   8  &  8 &   8   & 8 \\
\hline
Patience       & \{3, 10 \} & 4    &  9  & 10 & 8 &   8   & 9 \\
\hline
\end{tabular}%
}
\caption{Best hyperparameter values chosen for six generative models after 200 trials using Optuna \cite{optuna_2019}, a Bayesian optimization framework }
\label{tab:hyperparameters}
\end{table}
\section{Results}
\subsection{Odor Likeliness equation}
Table \ref{table:odor_likliness} highlights that the logistic regression model achieved strong performance with a ROC AUC score of 0.9701, indicating excellent discriminatory power of classifying molecules as odorous or odorless. The key metrics illustrated include recall (0.9448), highlighting its ability to identify true positives among all actual positives; precision (0.9121), signifying a high level of correctness in the positive predictions; accuracy (0.9253), reflecting the proportion of all correct predictions (both positive and negative) out of the total predictions; and an F1 Score (0.9282), balancing the trade-off between precision and recall, showcasing the model's robustness in handling imbalanced data. A heatmap of feature correlations indicated a minimum correlation of $-0.25$ between SlogP VSA3(Volsurf Surface Area Descriptor 3) and logP and a maximum of 0.72 between Molecular Weight and SlogP\_VSA3. The top five features after multicollinearity removal were logP, Molecular Weight, SlogP\_VSA3, Fraction of sp$^2$ Hybridized Atoms, and FCFP4 Count. Equation \ref{eq:odor_likeliness} is the logistic regression equation that we obtain.
\begin{table}[ht]
\centering
\resizebox{\textwidth}{!}{%
\begin{tabular}{|l|c|c|c|c|}
\hline
\textbf{Model} & \textbf{GDB-17 Criteria(\%)} \cite{ruddigkeit2012enumeration} & \textbf{Rule of Three(\%)} \cite{mayhew2022transport} & \textbf{Fragrance-Like (FL) Property} \cite{ruddigkeit2014expanding} & \textbf{Logistic Regression} \\
\hline
GAE & $100 \pm 0$ & $68.7 \pm 6.048$ & $46.02 \pm 8.96$  & $99.05 \pm 0.38$ \\
\hline
VGAE & $99.90 \pm 0.18$ &35.95 $\pm12.13 $  & $10.37 \pm2.49 $ & $ 97.17\pm1.13 $ \\
\hline
ARGA &  $ 100\pm0 $   & $ 82.88\pm1.79 $ & $ 81.91\pm3.78 $ & $100\pm0 $ \\
\hline
ARGVA & $84.39 \pm 0.17$ & $93.75 \pm0.22 $ & $93.66 \pm0.25 $  & $ 98.81\pm0.41 $ \\
\hline
Diffusion Model & $97.34 \pm 1.29$ & $93.28 \pm 2.12$  & $82.22 \pm 1.64$ & $99.98 \pm 0.02$  \\

\hline
Transformer model & $99.46 \pm 0.61$ & $57.04 \pm 1.71$  & $37.20 \pm 5.13$  & $99.46 \pm 0.31$  \\
\hline
\end{tabular}%
}
\caption{Evaluation of the percentage of fragrant molecules among novel generated molecules using  using the GDB-17 criteria, the Rule of three and Fragrance-Like (FL) Property with logistic regression \cite{ruddigkeit2012enumeration,mayhew2022transport,ruddigkeit2014expanding} }
\label{table:compare_odor_likliness}
\end{table}
\subsection{Benchmarking}

\begin{table}[tbp]
\centering
\resizebox{\textwidth}{!}{%
\begin{tabular}{|l|c|c|c|c|c|c|}
\hline
\textbf{Model} & \textbf{Validity} & \textbf{Uniqueness} & \textbf{Novelty} & \textbf{Diversity Score} & \textbf{Scaffold Similarity (Scaff)} & \textbf{Similarity to Nearest Neighbor (SNN)} \\
\hline
GAE & $0.94\pm0.08$ & $1\pm0$ & $0.86\pm0.04$  & $0.92\pm0.01$ & $0.52\pm0.02$ & $0.43\pm0.02$ \\
\hline
VGAE & $0.90\pm0.09$ & $1\pm0$ & $0.98\pm0.01$ & $0.92\pm0$ & $0.47\pm0.01$ & $0.34\pm0.02$ \\
\hline
ARGA & $1\pm0$ & $1\pm0$ & $0.92\pm0.01$ & $0.92\pm0.02$ & $0.49\pm0.01$ & $0.45\pm0.02$ \\
\hline
ARGVA & $0.99\pm0.01$ & $1\pm0$ & $0.99\pm0$ & $0.81\pm0$ & $0.37\pm0$ & $0.46\pm0$ \\
\hline
Diffusion model & $1\pm0$ & $0.98\pm0$ & $0.87\pm0.02$ & $0.89\pm0$ & $0.64\pm0.05$ & $0.47\pm0.02$ \\
\hline
Transformer model & $0.92\pm0.03$ & $1\pm0$ & $0.80\pm0.04$ & $0.93\pm0$ & $0.58\pm0.01$ & $0.43\pm0$ \\
\hline
\end{tabular}
}
\caption{Analysis of generative models on MOSES\cite{polykovskiy2020molecular} benchmarks}
\label{tab:fragrance_models}
\end{table}

Table \ref{table:compare_odor_likliness} presents a comparison of the percentage of novel fragrant molecules generated from the logistic regression equation and the other three models which propose a odor likeliness criteria based on different conditions. The FL property\cite{ruddigkeit2014expanding} and the criteria of the GDB-17 \cite{ruddigkeit2012enumeration}  do not pass molecules other than C, H, O, S, N and halogens, whereas the rule of three and the logistic regression offers an approach where it also includes molecules which do have these elements.  Each criterion applies a distinct set of filters. From the analysis in Table \ref{table:compare_odor_likliness}, it is evident that the logistic regression model identifies most of the generated molecules as odorous. This observation was further corroborated using a GNN-based odor prediction model, which confirmed that all molecules identified as odorous by the logistic regression also confirmed odor for the molecule in the GNN model \cite{lee2023principal}. Among the criteria, only GDB-17 showed comparable performance to logistic regression. This variation is mainly because criteria like GDB-17, Rule of three, and FL property primarily rely on molecular weight and heteroatom counts , whereas the logistic regression model assesses 51 features and selects the most relevant molecules. The use of explainable AI methods, such as SHAP, combined with the removal of features based on correlations, ensures that only the most significant features are retained in the logistic regression.

\begin{equation}
\begin{aligned}
\text{Odor\_Likeliness} &= -3.6592 \\
&\quad + (7.0771 \times \text{logP}) \\
&\quad + (-6.2811 \times \text{Molecular Weight}) \\
&\quad + (1.1403 \times \text{SlogP\_VSA3}) \\
&\quad + (0.5869 \times \text{Fraction of Sp}^2 \text{ Hybridized Atoms}) \\
&\quad + (1.9262 \times \text{FCFP4 Count})
\end{aligned}
\label{eq:odor_likeliness}
\end{equation}

\begin{table}[ht]
\centering
\begin{tabular}{|l|l|}
\hline
\textbf{Metric/Attribute} & \textbf{Value} \\
\hline
ROC AUC Score & 0.9701 \\
\hline
Recall & 0.9448 \\
\hline
Precision & 0.9121 \\
\hline
Accuracy & 0.9253 \\
\hline
F1 Score & 0.9282 \\
\hline
Original Number of Attributes & 51 \\
\hline
Attributes Removed Due to High Correlation (\textgreater 0.75) and VIF (\textgreater 5) & 31 \\
\hline
Attributes Available for Building the Equation & 20 \\
\hline
Attributes in Final Model Based on SHAP Values & 5 \\
\hline
\end{tabular}
\caption{Model evaluation metrics and attribute information for the logistic regression model.}
\label{table:odor_likliness}
\end{table}

\begin{figure}[htbp]
    \centering
        \begin{minipage}{0.9\textwidth}
            \centering
            \begin{subfigure}[b]{0.36\textwidth}
                \centering
                \fbox{\hspace{0.04cm} 
                    \includegraphics[scale=0.325]{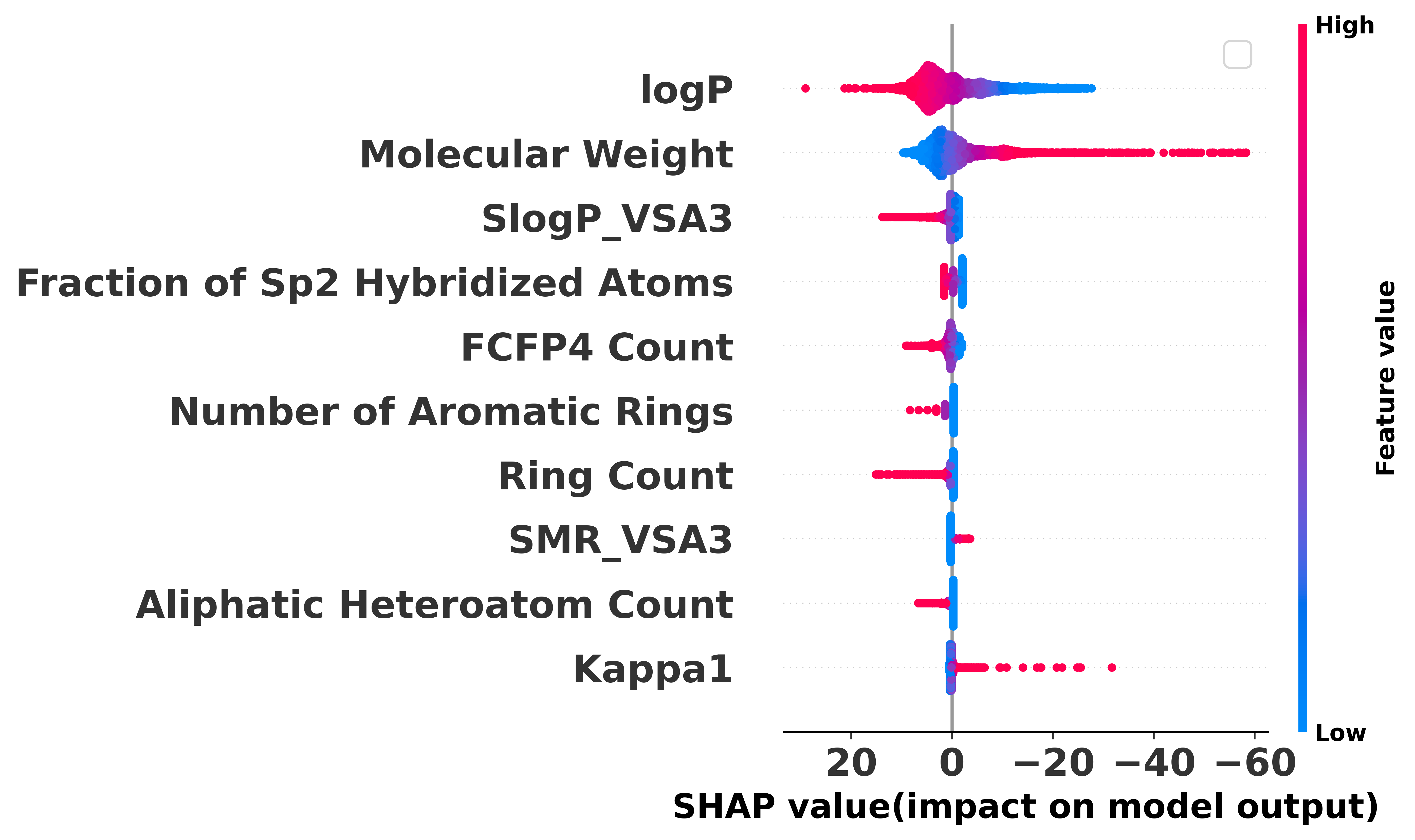}
                \hspace{0.04cm}}
                \caption{}
                \label{fig:shap}
            \end{subfigure}
            \hfill
            \begin{subfigure}[b]{0.36\textwidth}
                \centering
                \fbox{\hspace{0.04cm} 
                    \includegraphics[scale=0.225]{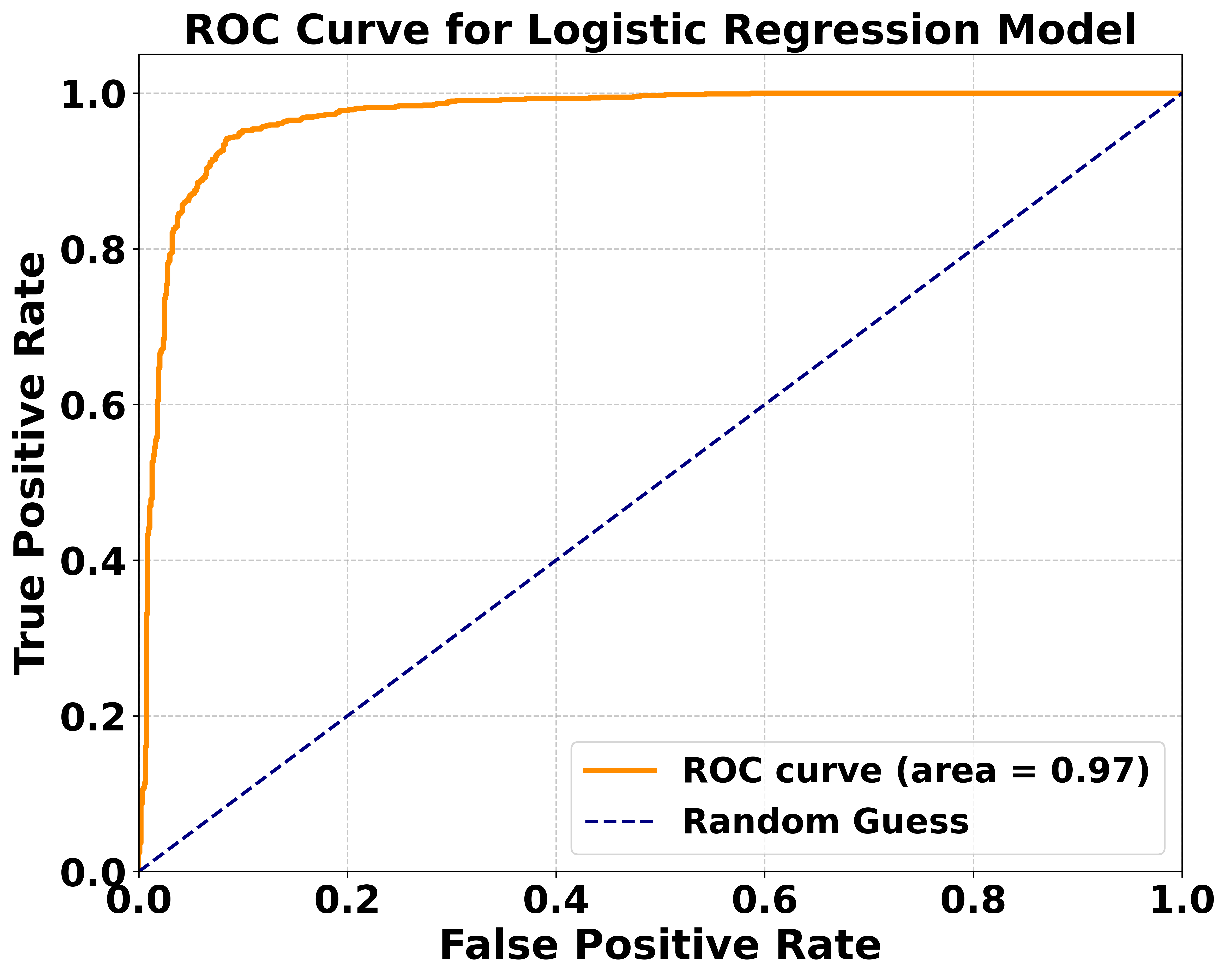}
                \hspace{0.04cm}}
                \caption{}
                \label{fig:auroc}
            \end{subfigure}
        \end{minipage}
    \caption{\small\textbf{Logistic Regression Model Performance Analysis.}  
    \textbf{(a) SHAP (SHapley Additive exPlanations) analysis:} Summary plot illustrating the contribution of key molecular descriptors to the prediction of fragrance likeliness.  
    \textbf{(b) Receiver Operating Characteristic (ROC) curve:} Demonstrates excellent performance with high true positive rates while maintaining low false positive rates.}
    \label{fig:model_analysis}
\end{figure}

The SHAP summary plot in Figure \ref{fig:shap} provides a view of the global importance and impact of molecular descriptors for predicting odor likelihood. Among the features, logP stands out as the most influential. Its high values positively contribute to odorous predictions and low values favor odorless classification. The higher values of molecular weight is associated with odorless predictions and lower values contribute to odorous predictions.  SlogP\_VSA3 demonstrates a complex relationship, where both the high and low values influence predictions in varying directions depending on their magnitude. Fraction of Sp² hybridized atoms contributes significantly to odorous predictions when its values are high. This reflects the importance of molecular planarity and aromaticity in predicting odor likelihood. FCFP4 count exhibits a strong positive contribution to odorous predictions when present in higher quantities, highlighting how functional group variety influences odor properties.  Other features, like ring count and aliphatic heteroatom count, show limited impact, with smaller SHAP value distributions. Due to their lesser impact we do not utilize them in the logistic regression equation. Collectively, the plot illustrates the importance of both structural and physicochemical properties in predicting odor likelihood, complementing instance-specific insights observed in the force plots\cite{lundberg2018explainable} and enhancing the model's interpretability.

 The benchmark from the MOSES paper \cite{polykovskiy2020molecular} was used to evaluate the models, revealing that while most models achieve high validity percentages, they vary significantly in diversity, similarity to the nearest neighbor (SNN), and scaffold similarity. Striking a balance between exploration and exploitation is essential i.e. a model that is good should not only generate a high number of valid molecules but also ensure structural diversity, both among the generated molecules and compared to the training samples.
 Among the models, ARGA, ARGVA, and the Diffusion model achieve perfect validity scores (1 ± 0) which indicates their robustness in generating chemically correct and stable structures. All models demonstrate near-perfect uniqueness, this is mainly because in the code we give a retry of 10 to ensure that all molecules are unique, though the Diffusion model shows a slightly lower score (0.98 ± 0.02). Novelty is a critical MOSES metric that evaluates the generation of molecules not present in the training set. ARGA, ARGVA, and the Diffusion model lead in this metric and it was observed that the Transformer model struggles with a lower score (0.80 ± 0.04) which suggests potential overfitting. Diversity, crucial for exploring broader chemical spaces is highest for the Transformer model (0.93 ± 0.02) but lowest for ARGVA (0.81 ± 0.02). This indicates varying capability of the models to generate structurally distinct molecules.
Scaffold similarity is highest for Diffusion model and Transformer model which signify their ability in exploring novel molecular backbones, with scores of 0.64 ± 0.05 and 0.58 ± 0.01 respectively. ARGA and ARGVA show conservative scaffold generation with the lowest scores (less than 0.37 ± 0.01). Finally, similarity to the nearest neighbor (SNN) reveals a balance between originality and resemblance to training data, where VGAE achieves the lowest SNN score (0.34 ± 0.02), demonstrating higher originality.  ARGVA has a higher value and generates molecules closer to the training set (0.46 ± 0). The results align well with MOSES’s objectives, showcasing the trade-offs inherent in molecular generative modeling. ARGA, ARGVA, and the Diffusion Model excel in validity and novelty, while the Transformer model leads in diversity and scaffold exploration, albeit at the expense of novelty. These findings highlight the strengths and limitations of each model, underscoring MOSES’s utility in evaluating and guiding the development of generative models tailored to specific molecular design goals. Future research should aim to improve the trade-offs between novelty, scaffold exploration, and diversity to push the boundaries of molecular generation.

\begin{figure}[htbp]
    \centering
    \fbox{ 
        \begin{minipage}[b]{\linewidth}
            \centering
            \begin{subfigure}{\linewidth}
                \centering
                \fbox{ 
                    \includegraphics[scale=0.25]{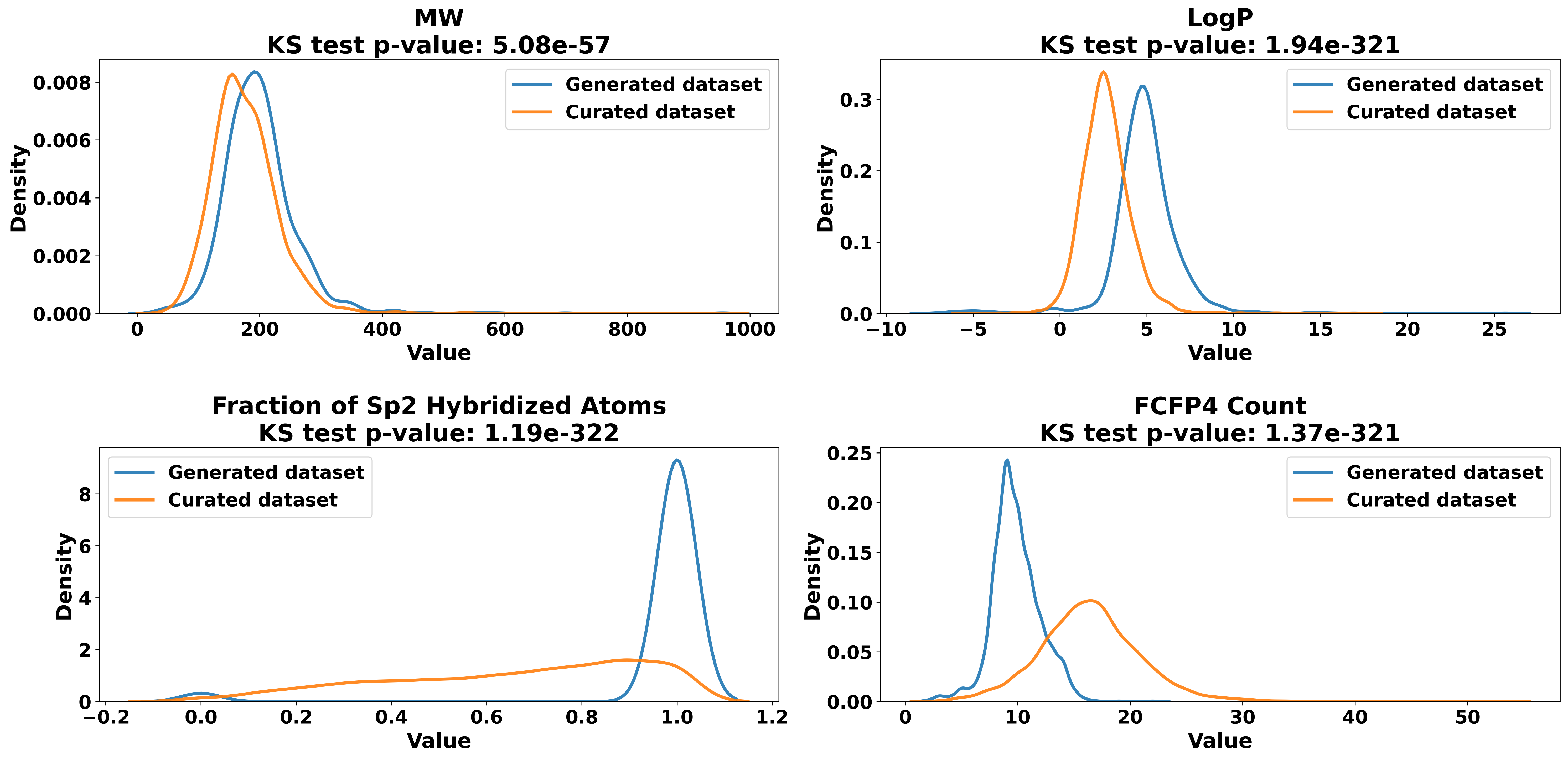}
                }
                \caption{}
                \label{fig:ks1_image}
            \end{subfigure}
            \begin{subfigure}{\linewidth}
                \centering
                \fbox{ 
                     \includegraphics[scale=0.24]{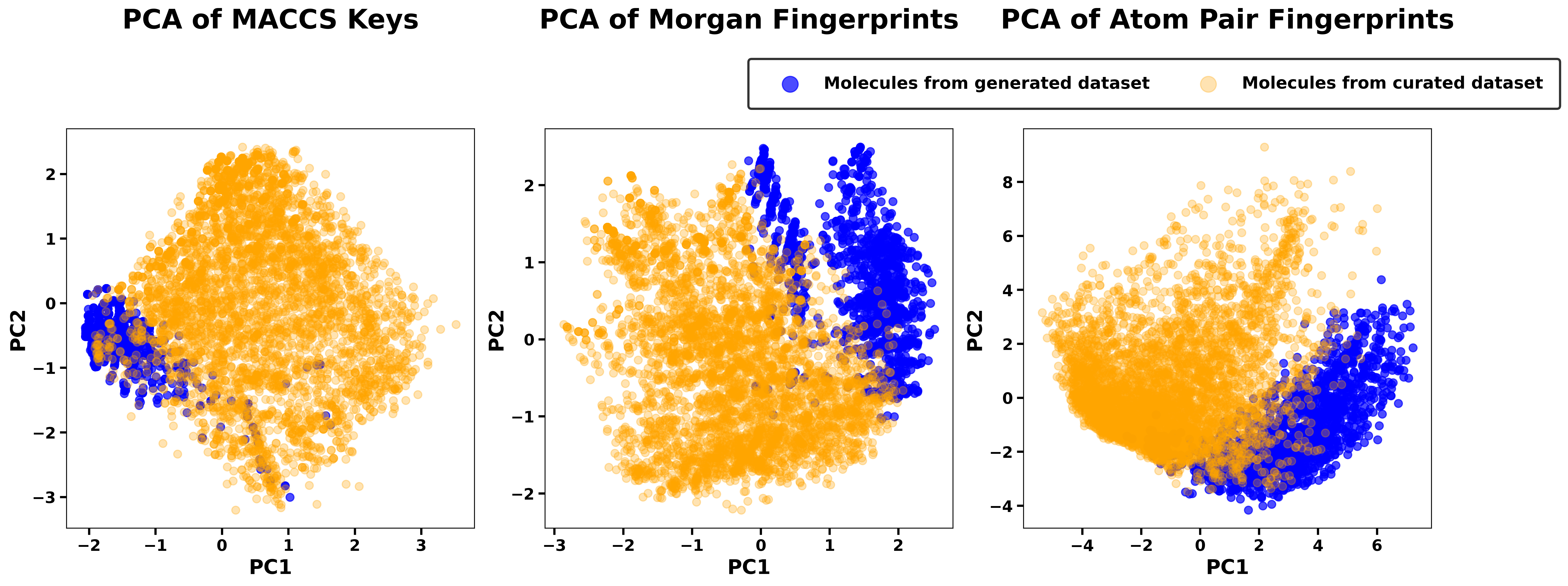}
                }
                \caption{}
                \label{fig:combined_fingerprints}
            \end{subfigure}
            \begin{subfigure}{\linewidth}
                \centering
                \fbox{ 
                     \includegraphics[scale=0.25]{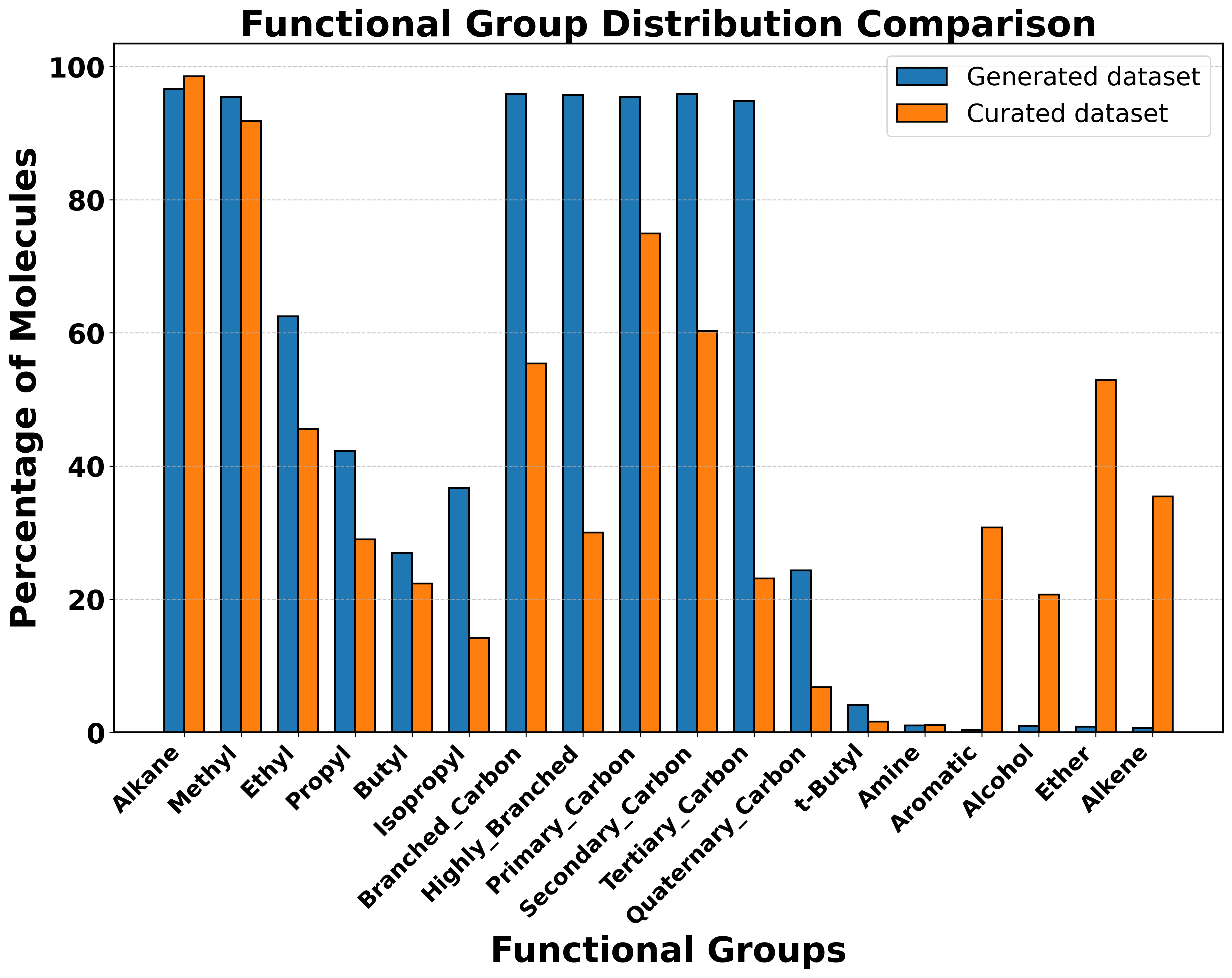}
                }
                \caption{}
                \label{fig:functional_groups}
            \end{subfigure}
        \end{minipage}
    }
    \caption{\small\textbf{Molecular properties comparison of generated molecules from ARGVA.} \textbf{(a)} KS test of parameters used for odor likeliness. \textbf{(b)} Analysis of the fingerprints. \textbf{(c)} Functional group analysis of the generated set. Refer the supplementary section for results of other models}
    \label{fig:main_figure}
\end{figure}

\section{Discussions}
\subsection{Benchmarking}
While it is observed that all models achieve high validity and uniqueness, trade-offs between novelty, diversity, and scaffold similarity persist. ARGA, ARGVA and Diffusion model excels in generating novel molecules outside the training set but exhibit limitations in scaffold innovation. However, their low scaffold similarity scores suggest limitations in creating innovative molecular backbones, which are often necessary for identifying unique and novel fragrance profiles. On the other hand, the Transformer model demonstrates strong diversity and scaffold similarity, though its lower novelty score indicates potential overfitting.  However, its relatively low novelty score points to overfitting tendencies, potentially limiting its capacity to explore completely new regions of the fragrance chemical space. This could pose challenges for applications requiring unique olfactory signatures. As per the benchmark results the Diffusion model stands out with a balanced performance across novelty, diversity, and scaffold similarity, making it ideal for scaffold exploration tasks. Such a balanced approach is essential for identifying molecules that are both novel and aligned with olfactory design goals. The comparison of ARGVA is highlighted in Figure \ref{fig:main_figure}, where Figure \ref{fig:ks1_image} shows the KS test results, Figure \ref{fig:combined_fingerprints} show the fingerprints and Figure \ref{fig:functional_groups} highlight the various functional groups in the generated dataset. Similar results for other models is available in the supplementary information.

\begin{figure}[htbp]
    \centering
    \fbox{%
        \begin{minipage}{0.98\textwidth}
            \centering
            \begin{minipage}{\textwidth}
                \centering
                \fbox{%
                    \begin{minipage}{0.95\textwidth}
                        \centering
                         \includegraphics[scale=0.455]{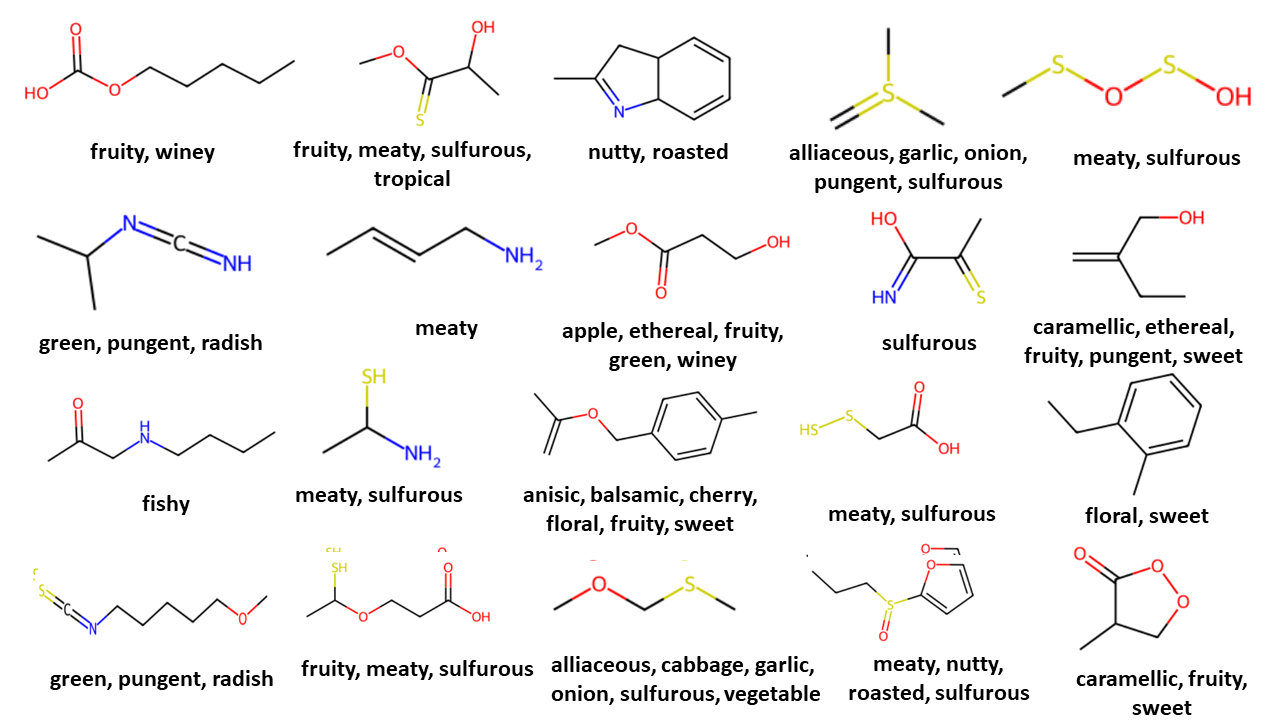}
                    \end{minipage}
                }
                \vspace{0.1em}
                \centerline{(a)}
                \label{fig:transformer_molecules}
            \end{minipage}
            \vspace{0.1em}
             \begin{minipage}{\textwidth}
                \centering
                \fbox{%
                    \begin{minipage}{0.95\textwidth}
                        \centering
                        \includegraphics[scale=0.455]{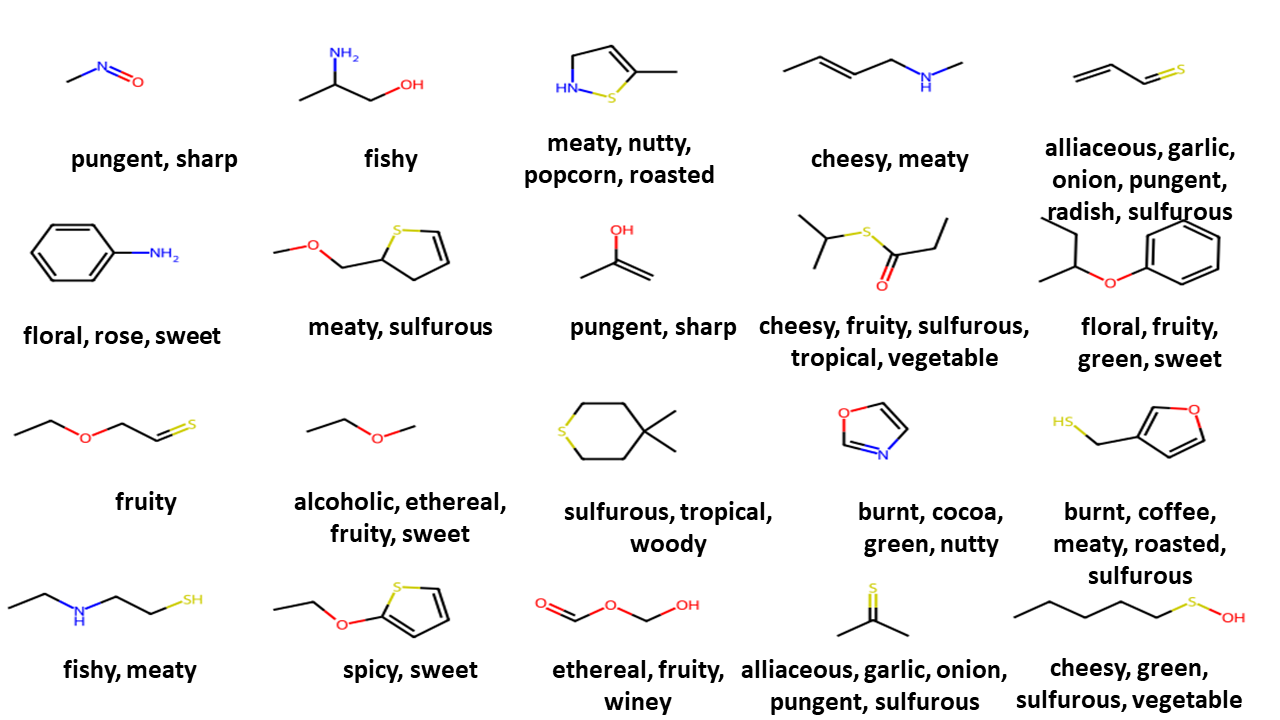}
                        \label{fig:gae_molecules}
                    \end{minipage}
                }
                \vspace{0.1em}
                \centerline{(b)}
            \end{minipage}
        \end{minipage}%
    }
    
    \vspace{0.1em}
    \caption{\small\textbf{Sample of novel generated molecules (which were not in the training set): Visualization of molecular structures and their predicted odors, assigned from 138 odor labels using graph neural networks.} \textbf{(a)} Results from transformer model \textbf{(b) }Results from GAE model. Refer to the supplementary section for results of other graph generative models.}
    \label{fig:generated_molecules}
\end{figure}
    
    
\begin{figure}[htbp]
    \centering
    \fbox{ 
        \begin{minipage}{0.95\textwidth} 
            \centering

            \begin{subfigure}[b]{0.48\textwidth} 
                \includegraphics[width=\textwidth]{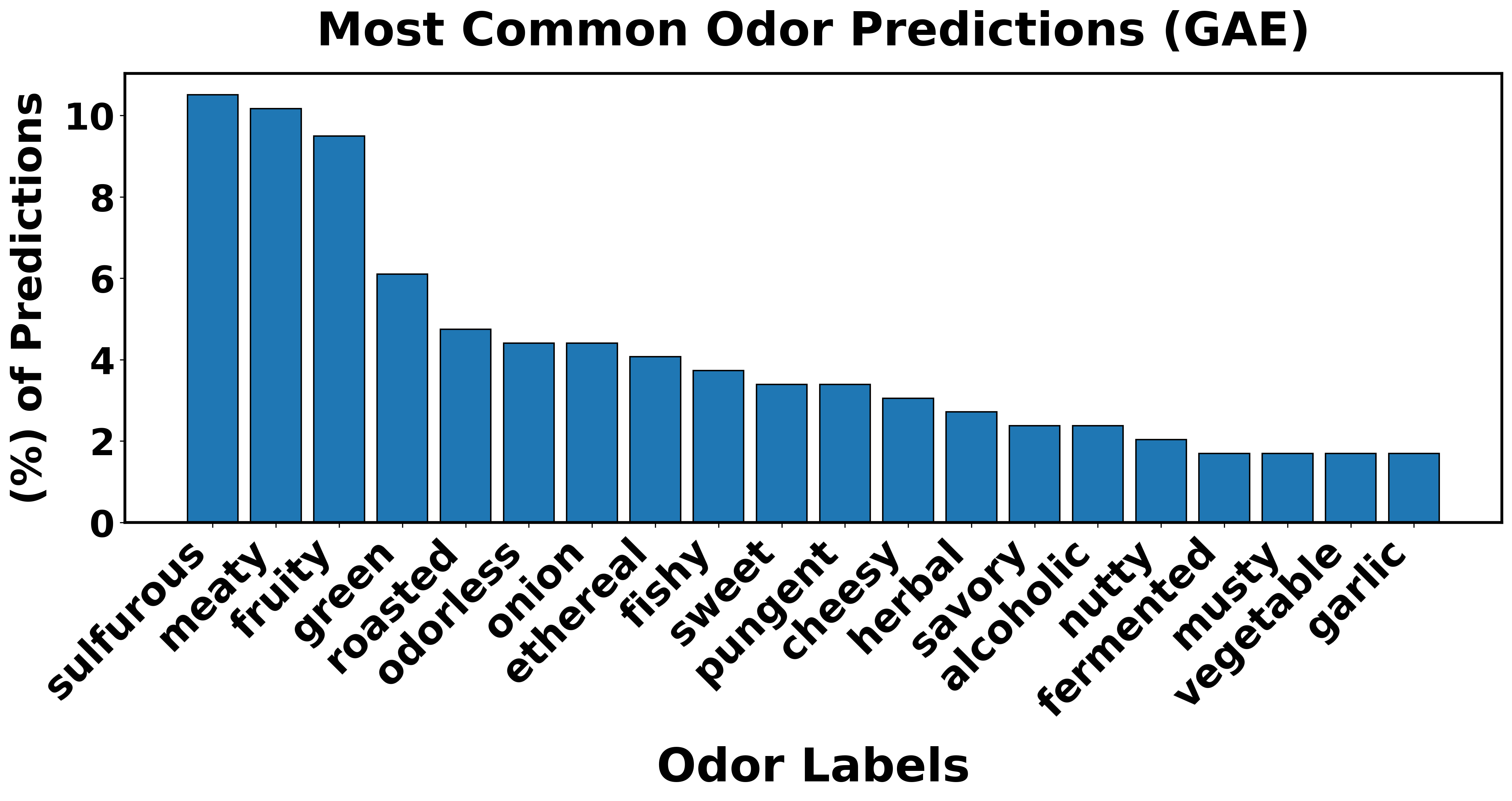}
                \caption{}
                \label{fig:top_odor_per_ARGA}
            \end{subfigure}
            \hspace{0.5em} 
            \begin{subfigure}[b]{0.48\textwidth} 
                \includegraphics[width=\textwidth]{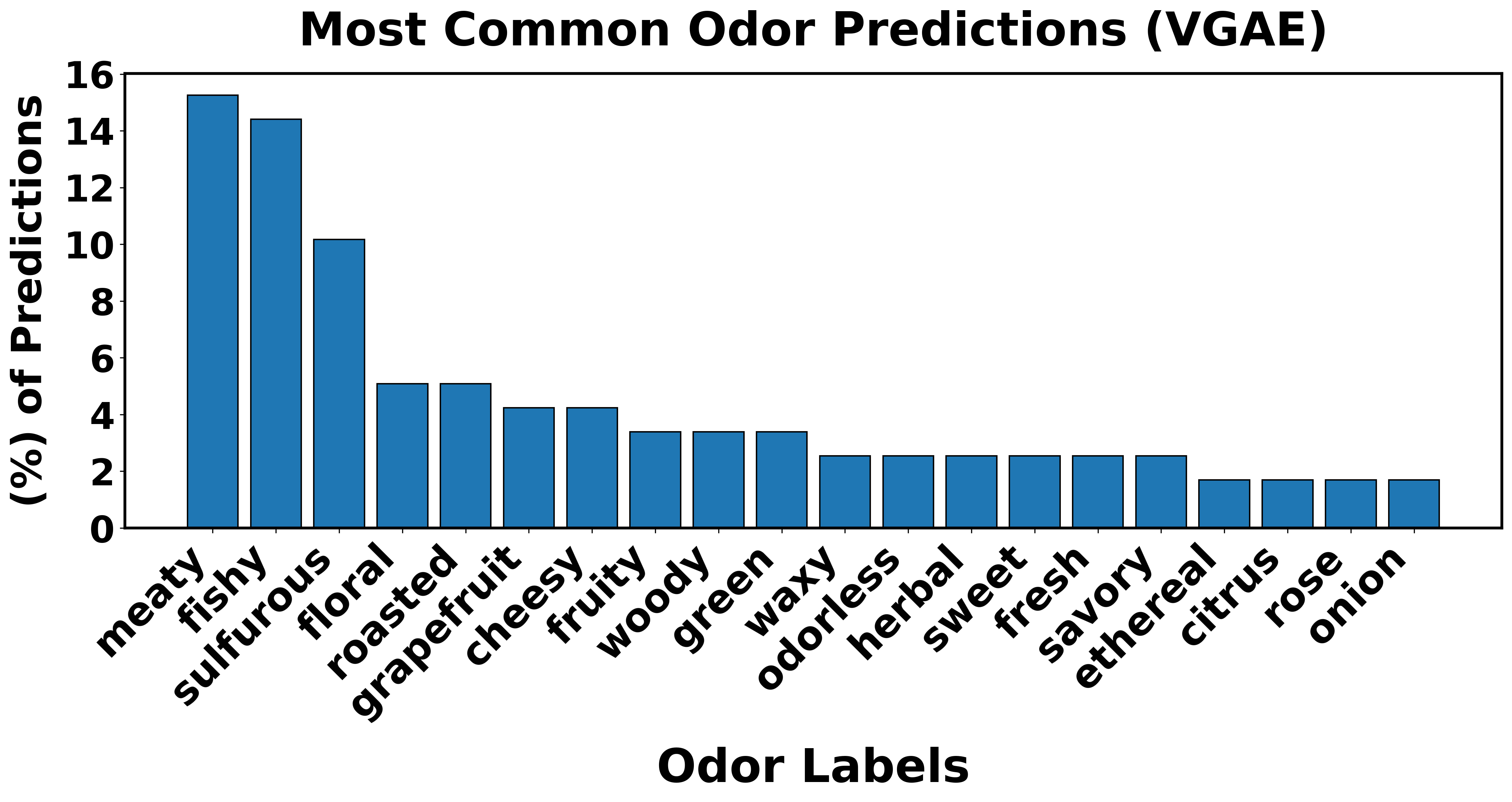}
                \caption{}
                \label{fig:top_odor_per_ARGVA}
            \end{subfigure}

            \vspace{1em} 

            \begin{subfigure}[b]{0.48\textwidth}
                \includegraphics[width=\textwidth]{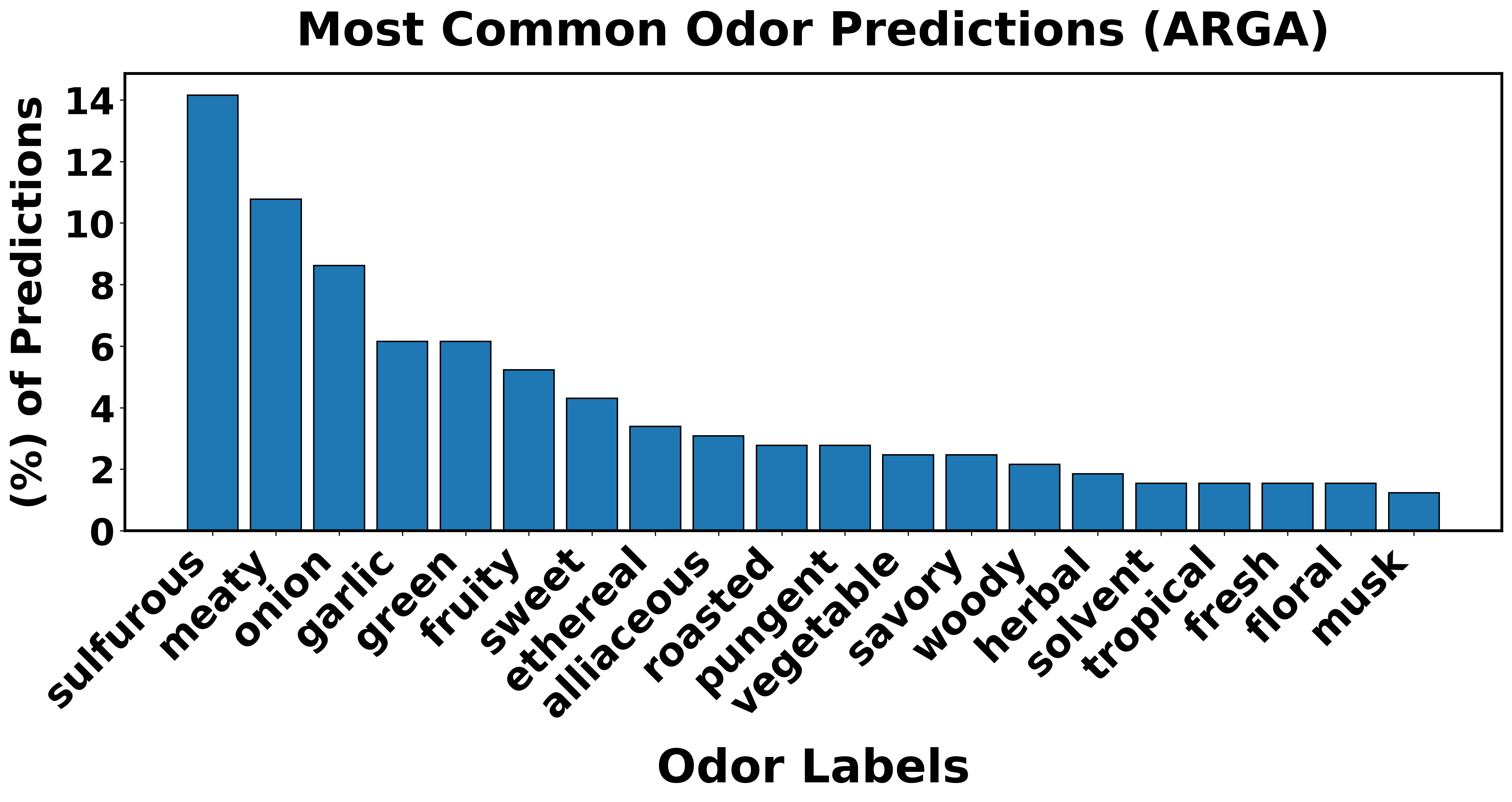}
                \caption{}
                \label{fig:top_odor_per_Diffusion}
            \end{subfigure}
            \hspace{0.5em} 
            \begin{subfigure}[b]{0.48\textwidth}
                \includegraphics[width=\textwidth]{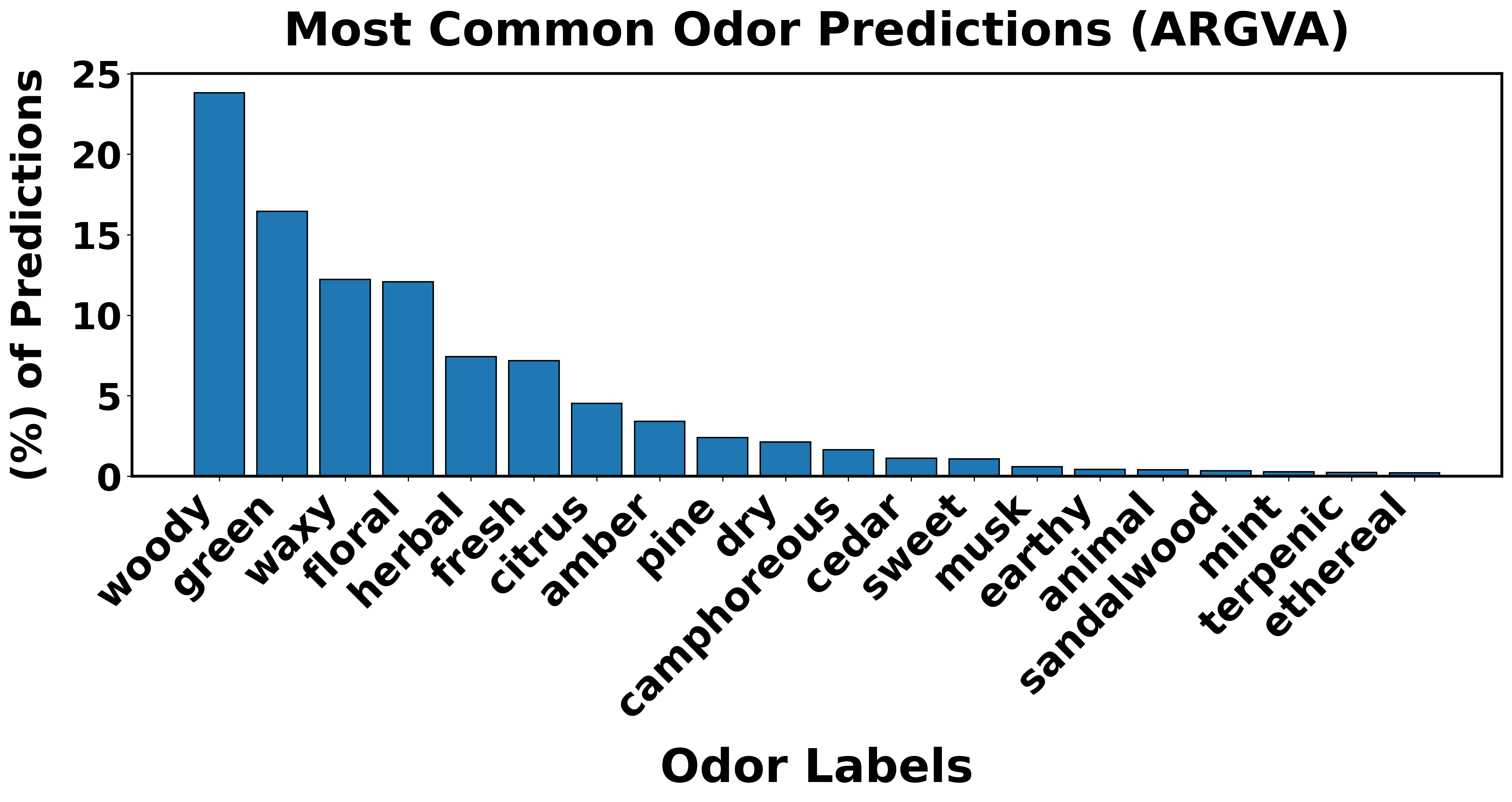}
                \caption{}
                \label{fig:top_odor_per_Transformer}
            \end{subfigure}

            \vspace{1em} 

            \begin{subfigure}[b]{0.48\textwidth}
                \includegraphics[width=\textwidth]{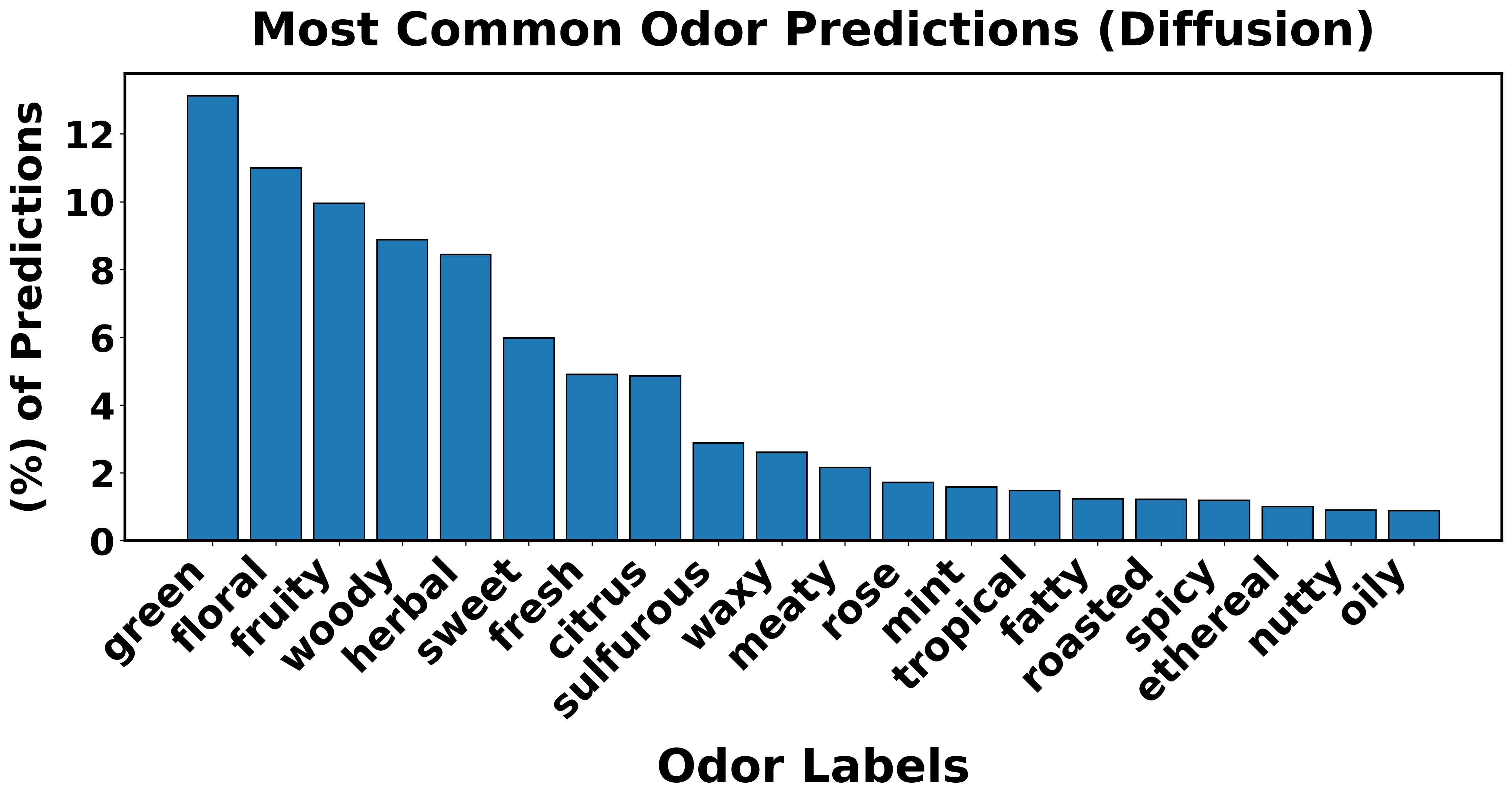}
                \caption{}
                \label{fig:top_odor_per_Model5}
            \end{subfigure}
            \hspace{0.5em} 
            \begin{subfigure}[b]{0.48\textwidth}
                \includegraphics[width=\textwidth]{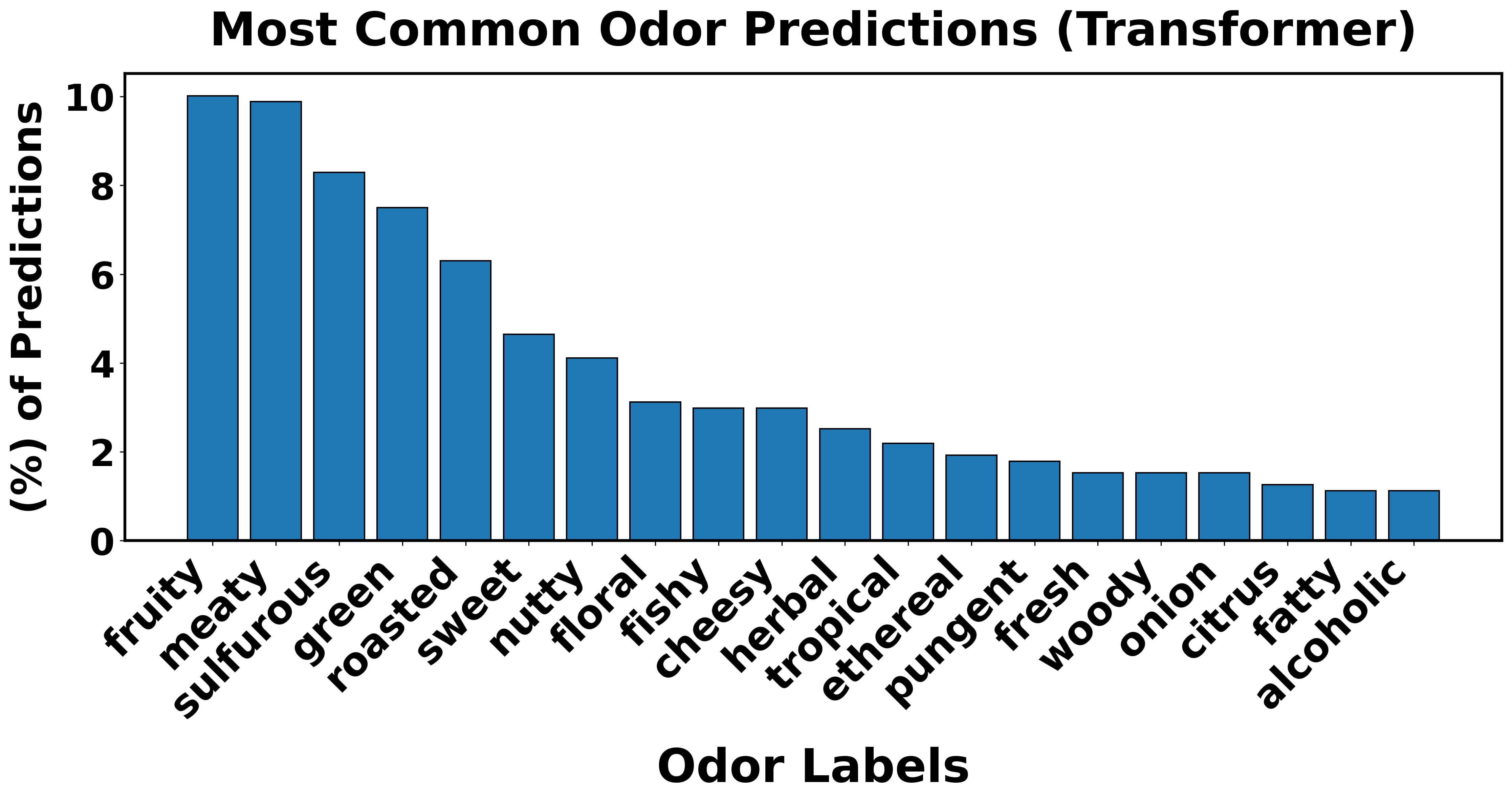}
                \caption{}
                \label{fig:top_odor_per_Model6}
            \end{subfigure}
             
            
        \end{minipage}
    }
    \caption{\small\textbf{Comparison of common odors predicted by various generative models:} It can be observed that whereas some models produces variety others produce same odors in high quantity. Odor labels predicted for molecules generated by \textbf{(a)} GAE \textbf{(b)} VGAE \textbf{(c)} ARGA \textbf{(d)} ARGVA  }
\label{fig:top_odors_comparison}
\end{figure}
\begin{figure}[htbp]
    \centering
    
    \fbox{
        \begin{minipage}{1\textwidth}
            \centering
            
            \fbox{
                \begin{minipage}{0.95\textwidth}
                    \centering
                    \begin{subfigure}[b]{1\textwidth}
                        \includegraphics[width=\linewidth]{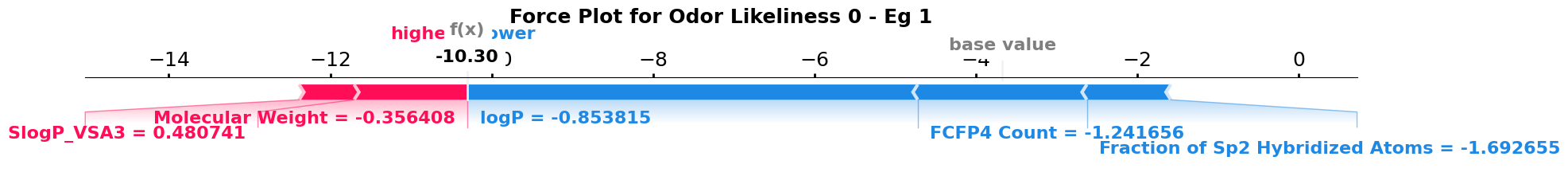}

                        \caption{Example 1 - Odor Likeliness 0 (Odorless)} \label{fig:force_plot_class_0_example_1}
                    \end{subfigure}
                    \vspace{0.5cm}  
                    \begin{subfigure}[b]{1\textwidth}
                        \includegraphics[width=\linewidth]{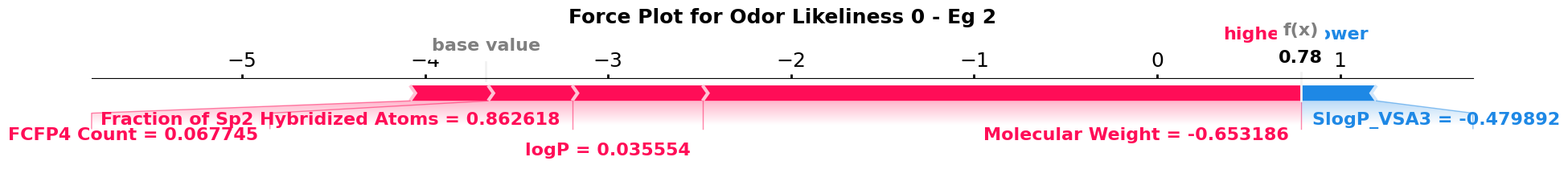}
                        \caption{Example 2 - Odor Likeliness 0 (odorless)} \label{fig:force_plot_class_0_example_2}
                    \end{subfigure}
                    \vspace{0.5cm}  
                    \begin{subfigure}[b]{1\textwidth}
                        \includegraphics[width=\linewidth]{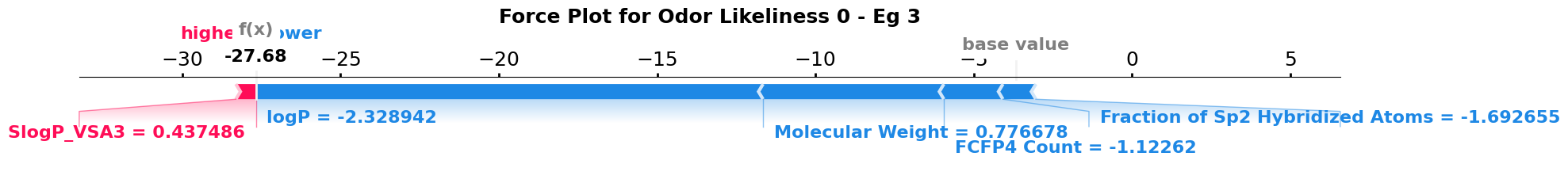}
                        \caption{Example 3 - Odor Likeliness 0 (odorless)} \label{fig:force_plot_class_0_example_3}
                    \end{subfigure}
                \end{minipage}
            }
            
            \vspace{1cm}  

            \fbox{
                \begin{minipage}{0.95\textwidth}
                    \centering
                    \begin{subfigure}[b]{1\textwidth}
                        \includegraphics[width=\linewidth]{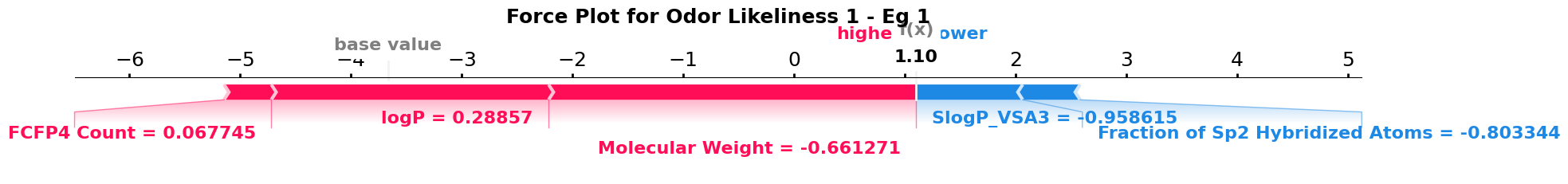}
                        \caption{Example 1 - Odor Likeliness 1 (odorous)} \label{fig:force_plot_class_1_example_1}
                    \end{subfigure}
                    \vspace{0.5cm}  
                    \begin{subfigure}[b]{1\textwidth}
                        \includegraphics[width=\linewidth]{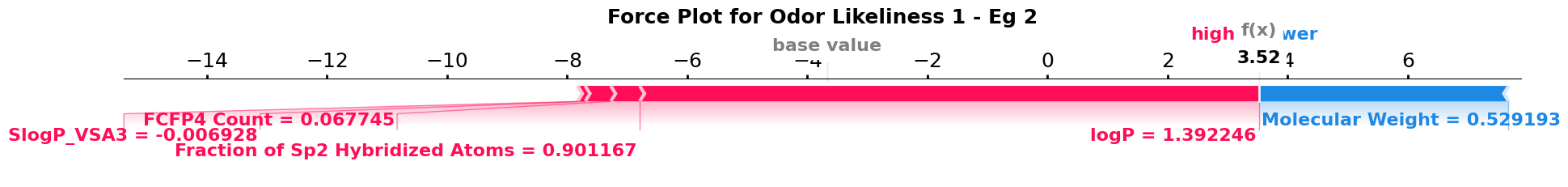}
                        \caption{Example 2 - Odor Likeliness 1 (odorous)} \label{fig:force_plot_class_1_example_2}
                    \end{subfigure}
                    \vspace{0.5cm}  
                    \begin{subfigure}[b]{1\textwidth}
                        \includegraphics[width=\linewidth]{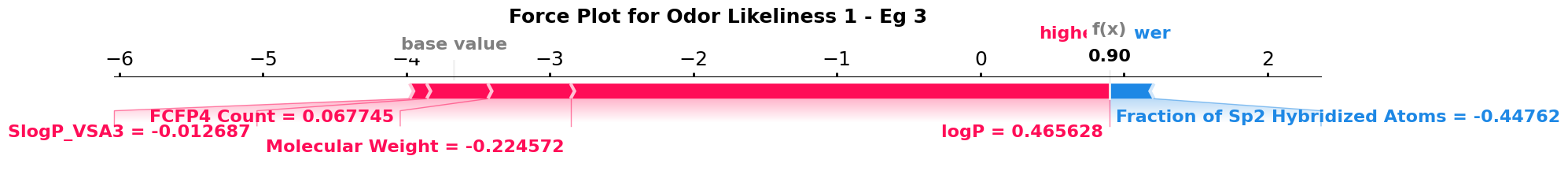}
                        \caption{Example 3 - Odor Likeliness 1 (odorous)} \label{fig:force_plot_class_1_example_3}
                    \end{subfigure}
                \end{minipage}
            }
        \end{minipage}
    }

    \caption{\small\textbf{SHAP Force Plots for Odor Likeliness Prediction:} Visualizing the contribution of top features for odorless (class 0: Odor Likeliness = 0) and odorous (class 1: Odor Likeliness = 1) molecules. Each plot illustrates how individual feature values impact the predicted odor likeliness, with SHAP values showing the magnitude and direction of their influence} \label{fig:force_plots}
\end{figure}

\subsection{Odor label prediction}
The molecular substructures associated with odor descriptors often overlap as similar fragments can correspond to different perceptual descriptors\cite{sell2006unpredictability}. The structure odor relationship is complex and unpredictable. Smell characteristics generally arise from a mosaic of overlapping features spanning multiple odor classes. Similarly, molecular substructures associated with odor descriptors often overlap, with similar fragments linked to different perceptual descriptors. This is exactly what happens in molecule-receptor interactions, where parts of a single molecule activate multiple receptors.
The training dataset’s odor category distribution which is  dominated by \desc{fruity}, \desc{green}, and \desc{sweet} labels influences the model predictions. Figure \ref{fig:top_odors_comparison} highlights the odor labels generated by different generative models. It is evident from the figure \ref{fig:top_odor_per_Model5} and figure \ref{fig:top_odor_per_Transformer} that Diffusion and Transformer models frequently generate these categories due to high scaffold similarity (Scaff) scores (Transformer: 0.58 ± 0.01; Diffusion: 0.64 ± 0.05). Despite this bias, the Diffusion model effectively balances diversity and structural fidelity, while models like Transformer (diversity score: 0.93 ± 0.0) and GAE (0.92 ± 0.01) explore broader chemical spaces. Figure \ref{fig:generated_molecules} presents the molecular structures of the molecules generated by GAE and Transformer models, along with their predicted odors.

 In contrast, ARGVA amplifies dominating categories, possibly under-representing uncommon odors, with \desc{fruity} notes accounting for 23\% of its predictions as opposed to 7.9\% in the training data. In contrast, Diffusion and Transformer models produce more balanced distributions and maintain rarer categories like \desc{nutty}, \desc{citrus} and \desc{tropical}, which are critical for diverse applications such as fragrance design. ARGA generates diverse profiles by balancing dominant and less common odors, while GAE and VGAE exhibit a bias toward \desc{sulfurous} labels due to the presence of sulfur-containing functional groups.
Each model serves specific purposes. ARGVA excels at generating dominant odor profiles, while Diffusion and Transformer models balance frequent and rare odors. By combining diversity and structural fidelity it is observed that Diffusion and ARGA show promise for diverse olfactory applications, bridging data-driven predictions with practical chemical design.

\subsection{Analysis of Odor likeliness criteria}
Figure \ref{fig:force_plot_class_0_example_1}, \ref{fig:force_plot_class_0_example_2}, \ref{fig:force_plot_class_0_example_3}  and Figure \ref{fig:force_plot_class_1_example_1}, \ref{fig:force_plot_class_1_example_2}, \ref{fig:force_plot_class_1_example_3} show how the features contribute when the molecule is odorless(class=0) or odorous(class=1) for three examples, for  each class of the dataset. It is evident that the five important molecular descriptors encompass all the characteristics that influence a molecule's odor potential. LogP assesses lipophilicity, with higher values suggesting more hydrophobicity, which is required for odorant molecules to volatilize and successfully interact with olfactory receptors. Lower molecular weights generally correlate with higher fragrance likelihood. Generally molecules in the 30--300 Da range are optimal\cite{mayhew2022transport}, as larger ones often fail to reach olfactory receptors due to low volatility. Molecular weight accentuates the relationship between size and volatility. SlogP\_VSA3 assesses the total hydrophobic surface area, suggesting that molecules with greater hydrophobic regions interact more effectively with the lipid-rich olfactory membranes, thereby increasing odor detection. The fraction of sp² hybridized atoms explains the presence of aromatic structures and unsaturation, and its higher values are generally linked to increased fragrance potential.  Lastly, the FCFP4 count acts as a molecular fingerprint that records the variety of structural elements and functional groups. It represents 3D molecular characteristics through circular substructures around each atom in a 2D format. Thus we can say that higher counts of fragrance-relevant substructures enhance the likelihood of a molecule being perceived as odorous. Collectively, these descriptors highlight the intricate interplay of volatility, hydrophobicity, molecular structure, and functional groups in determining a molecule’s olfactory properties, making the logistic regression model an effective tool for predicting fragrance likeliness.  Only molecules identified as odorous are forwarded to odor prediction. This conserves resources by avoiding the processing of molecules known to be odorless and also minimizes the risk of incorrectly predicting odor for odorless molecules.
\section{Conclusion}
This study leverages generative models and logistic regression to navigate the fragrance space efficiently. It combines molecular generation, rigorous validation and odor prediction in one integrated framework. The present study also emphasizes the critical role of key physicochemical features in predicting fragrance likeliness and assigning odor labels. Our approach offers a valuable solution to handle the impracticality of obtaining odor labels through human evaluation and the scarcity of labeled fragrant datasets. The inclusion of diverse graph-based generative techniques which  ensure high validity, novelty, and structural diversity, align well with olfactory design goals. The main emphasis is on model interpretability and the code and dataset is available on the Github repository. This study will enable cost-effective and scalable exploration of the chemical space. This framework sets a foundation for advanced applications in olfactory research, practical chemical design, and the development of novel and diverse fragrance molecules.
\section{Associated Content}
The source code is available at \url{https://github.com/CSIO-FPIL/generative-odor}. The Python scripts of charts and other forms of analysis can also be found on this GitHub repository. the website developed for this work is available at \url{https://kumars8494.github.io/Exploring_Fragrance_Space_with_Generative_model/}
\subsection{Supporting Information}
 Table of features used to make the logistic regression equation; Logistic Regression Model Analysis: Heatmap of Attributes; Table of comparison of SHAP and VIF Values of features; Molecular properties comparison of generated molecules from GAE; Molecular properties comparison of generated molecules from VGAE; Sample of generated molecules along with odor labels predicted for VGAE; Molecular properties comparison of generated molecules from ARGA; Sample of generated molecules along with odor labels predicted for ARGA; Sample of generated molecules along with odor labels predicted for ARGVA; Sample of generated molecules along with odor labels predicted for ARGVA; Molecular properties comparison of generated molecules from Diffusion; Sample of generated molecules along with odor labels predicted for Diffusion model; Molecular properties comparison of generated molecules from Transformer; .
 \subsection{Acknowledgements}
We acknowledge Dr. Michael Schmuker for his constructive discussions, which significantly contributed to the advancement of this work.
\bibliography{achemso-demo}
\newpage


\begin{center}
    \LARGE \textbf{Supplementary Information: Navigating the Fragrance space Via Graph Generative Models And Predicting Odors}
\end{center}


\renewcommand{\thefigure}{S\arabic{figure}}
\setcounter{figure}{0}  

\begin{figure}[htbp]
    \centering
    \includegraphics[scale=0.45]{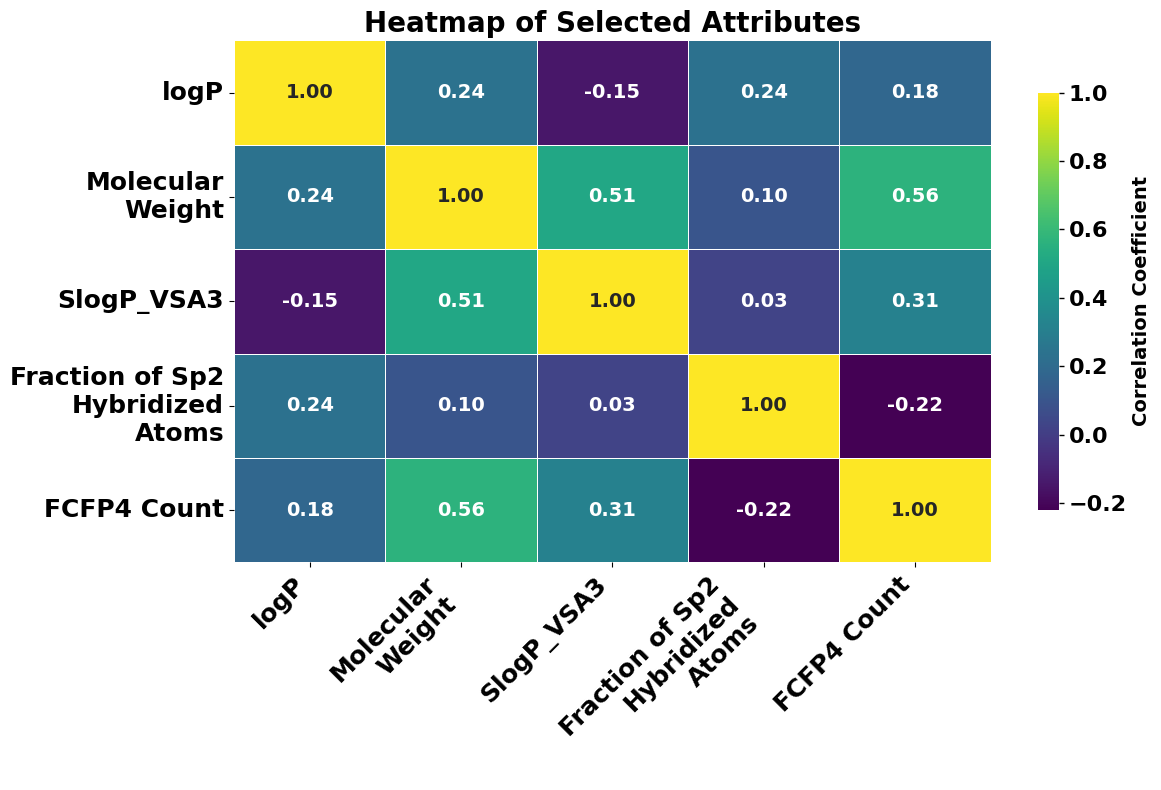}  

      \caption[]{Logistic Regression Model Analysis:Heatmap of Attributes}
    \label{fig:S1}
\end{figure}

\begin{longtable}{|p{2cm}|p{8cm}|}
\captionsetup{labelformat=empty}
\caption{Table S1: Features used to make the logistic regression equation} \\
\hline
\textbf{SR No.} & \textbf{Feature Name} \\
\hline
\endfirsthead

\hline
\textbf{SR No.} & \textbf{Feature Name} \\
\hline
\endhead

\hline
\endfoot

\hline
\endlastfoot

1 & QED \\
\hline
2 & logP \\
\hline
3 & Molecular Weight \\
\hline
4 & Number of Heavy Atoms \\
\hline
5 & TPSA \\
\hline
6 & Rotatable Bonds \\
\hline
7 & H-bond Donors \\
\hline
8 & H-bond Acceptors \\
\hline
9 & Ring Count \\
\hline
10 & Formal Charge \\
\hline
11 & Fraction of Sp2 Hybridized Atoms \\
\hline
12 & Number of Aromatic Rings \\
\hline
13 & Molar Refractivity \\
\hline
14 & Rotatable Bond Count \\
\hline
15 & Number of Heteroatoms \\
\hline
16 & Ipc \\
\hline
17 & Kappa1 \\
\hline
18 & Kappa2 \\
\hline
19 & Kappa3 \\
\hline
20 & LabuteASA \\
\hline
21 & PEOE\_VSA1 \\
\hline
22 & PEOE\_VSA2 \\
\hline
23 & PEOE\_VSA3 \\
\hline
24 & SMR\_VSA1 \\
\hline
25 & SMR\_VSA2 \\
\hline
26 & SMR\_VSA3 \\
\hline
27 & SlogP\_VSA1 \\
\hline
28 & SlogP\_VSA2 \\
\hline
29 & SlogP\_VSA3 \\
\hline
30 & FCFP4 Count \\
\hline
31 & ECFP4 Count \\
\hline
32 & Num Bridgehead Atoms \\
\hline
33 & Num Spiro Atoms \\
\hline
34 & Num Macrocycles \\
\hline
35 & Fsp3 \\
\hline
36 & Ring Atom Count \\
\hline
37 & Ring Bond Count \\
\hline
38 & Aliphatic Ring Count \\
\hline
39 & Aliphatic Heteroatom Count \\
\hline
40 & Aromatic Heteroatom Count \\
\hline
41 & Saturated Ring Count \\
\hline
42 & Aromatic Ring Count \\
\hline
43 & Hetero-Aliphatic Ring Count \\
\hline
44 & Hetero-Aromatic Ring Count \\
\hline
45 & Hetero-Saturated Ring Count \\
\hline
46 & Largest Ring Size \\
\hline
47 & Wildman-Crippen MR \\
\hline
48 & Average Molecular Weight \\
\hline
49 & Exact Molecular Weight \\
\hline
50 & Atom Pair Fingerprint Count \\
\hline
51 & Natural Product Likeness Score \\
\hline
\end{longtable}

\begin{table}[ht]
\captionsetup{labelformat=empty}
\caption{Table S2: Comparison of SHAP and VIF Values for Features} 
\centering
\resizebox{\textwidth}{!}{
\begin{tabular}{|l|c|l|c|}
\hline
\multicolumn{2}{|c|}{\textbf{SHAP Values}} & \multicolumn{2}{c|}{\textbf{VIF Values}} \\
\hline
\textbf{Feature} & \textbf{SHAP Value} & \textbf{Feature} & \textbf{VIF} \\
\hline
logP & 5.2019 & Average Molecular Weight & 0.085286 \\
\hline
Molecular Weight & 4.1472 & QED & 0.098205 \\
\hline
SlogP\_VSA3 & 1.0310 & FCFP4 Count & 0.214786 \\
\hline
Fraction of Sp2 Hybridized Atoms & 0.9714 & Fraction of Sp2 Hybridized Atoms & 0.263798 \\
\hline
FCFP4 Count & 0.8394 & Molecular Weight & 0.299668 \\
\hline
Number of Aromatic Rings & 0.7090 & Rotatable Bonds & 0.523401 \\
\hline
Ring Count & 0.4479 & SlogP\_VSA3 & 0.546333 \\
\hline
SMR\_VSA3 & 0.3924 & PEOE\_VSA2 & 0.726521 \\
\hline
Aliphatic Heteroatom Count & 0.3897 & Number of Aromatic Rings & 0.791741 \\
\hline
Kappa1 & 0.3195 & PEOE\_VSA3 & 0.817641 \\
\hline
Average Molecular Weight & 0.2136 & Kappa1 & 0.825312 \\
\hline
Rotatable Bonds & 0.2068 & Ring Count & 0.873277 \\
\hline
PEOE\_VSA2 & 0.1111 & SMR\_VSA3 & 0.897198 \\
\hline
PEOE\_VSA3 & 0.0722 & Aliphatic Heteroatom Count & 0.909384 \\
\hline
Num Spiro Atoms & 0.0301 & SMR\_VSA2 & 0.991069 \\
\hline
QED & 0.0271 & Num Spiro Atoms & 0.991984 \\
\hline
SMR\_VSA2 & 0.0219 & Kappa3 & 0.998421 \\
\hline
Formal Charge & 0.0115 & Formal Charge & 0.999686 \\
\hline
Kappa3 & 0.0022 & logP & 0.999796 \\
\hline
Ipc & 0.0005 & Ipc & 1.237964 \\
\hline
\end{tabular}
}

\label{tab:shap_vif_comparison}
\end{table}

\begin{figure}[htbp]
    \centering
    \fbox{ 
        \begin{minipage}[b]{\linewidth}
            \centering
            \begin{subfigure}{\linewidth}
                \centering
                \fbox{ 
                    \includegraphics[scale=0.27]{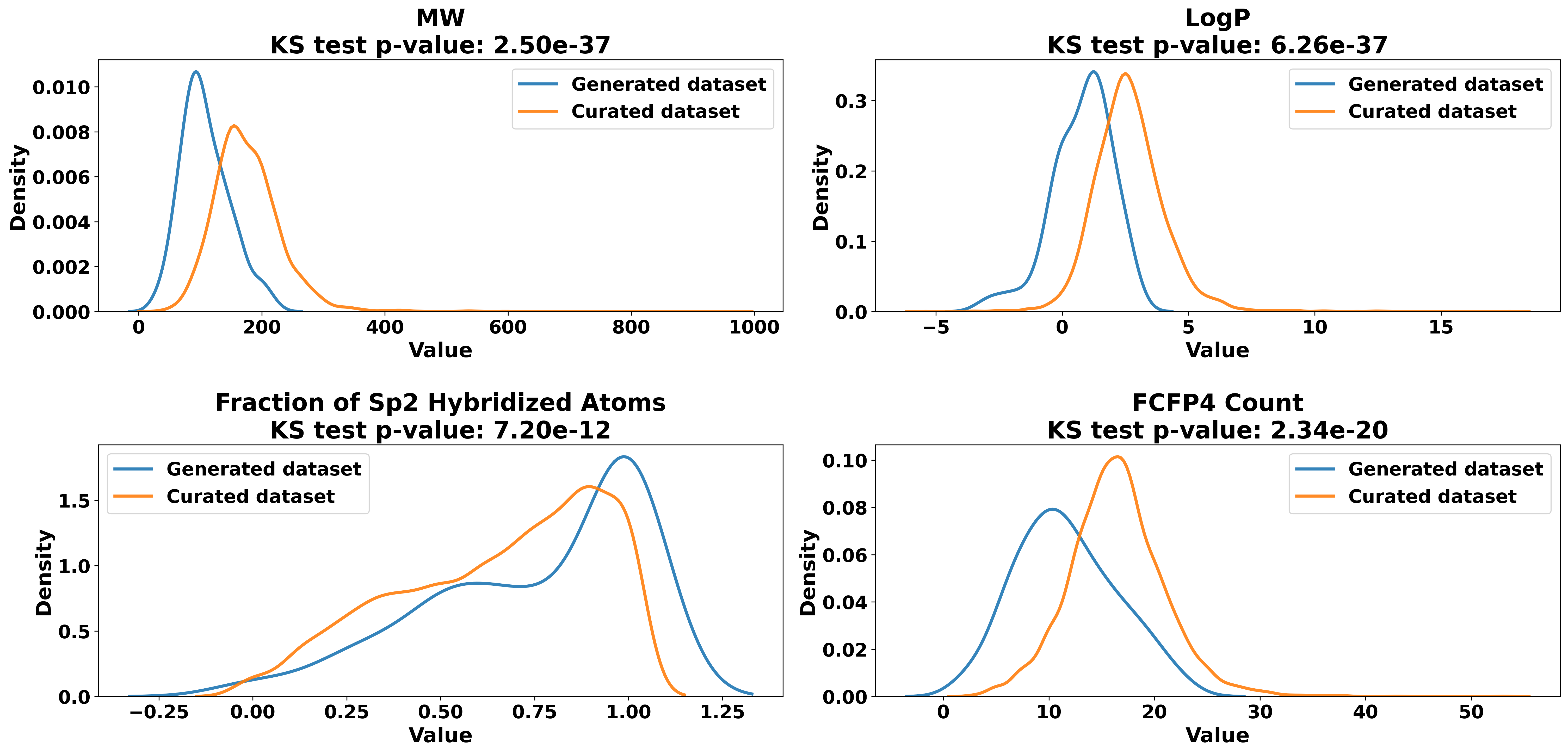}
                }
                \caption{}
                \label{fig:ks11_image}
            \end{subfigure}

            \vspace{1em} 

            \begin{subfigure}{\linewidth}
                \centering
                \fbox{ 
                     \includegraphics[scale=0.27]{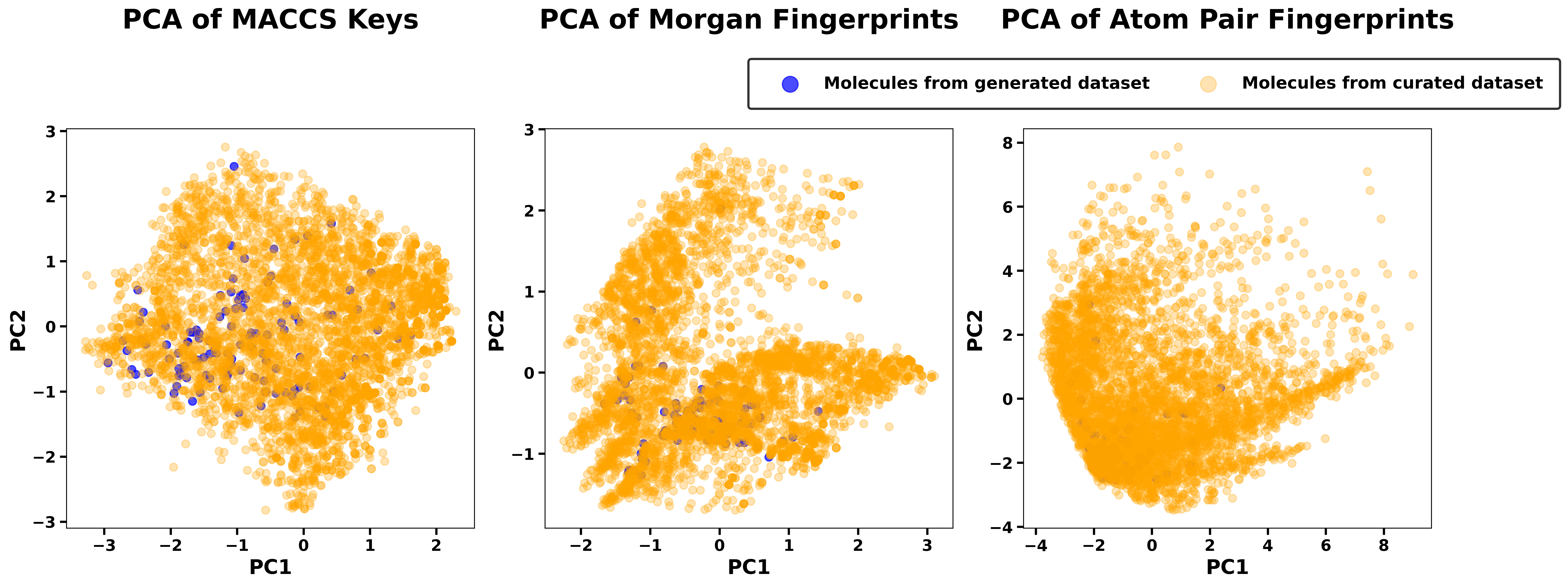}
                }
                \caption{}
                \label{fig:combinedf_fingerprints}
            \end{subfigure}
            \vspace{0.5em} 

            \begin{subfigure}{\linewidth}
                \centering
                \fbox{ 
                     \includegraphics[scale=0.20]{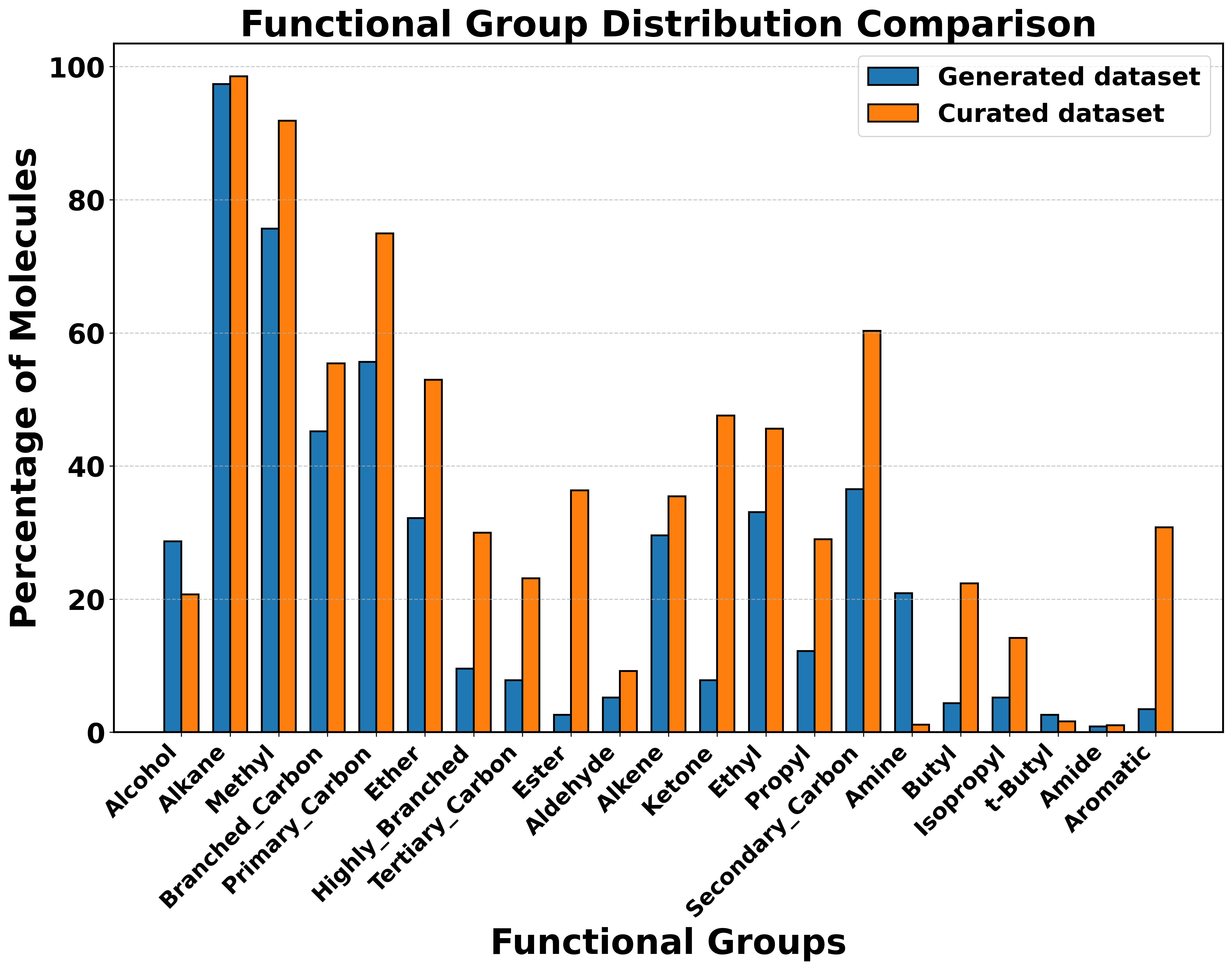}
                }
                \caption{}
                \label{fig:functional_groupes}
            \end{subfigure}
        \end{minipage}
    }
    
      \caption{\small\textbf{Molecular properties comparison of generated molecules from GAE.} (a) KS test of parameters used for odor likeliness. (b) Analysis of the fingerprints. (c) Functional group analysis of the generated set.}
    \label{fig:main_figure_g}

\end{figure}
    

\begin{figure}[htbp]
    \centering
    \fbox{ 
        \begin{minipage}[b]{\linewidth}
            \centering
            \begin{subfigure}{\linewidth}
                \centering
                \fbox{ 
                    \includegraphics[scale=0.27]{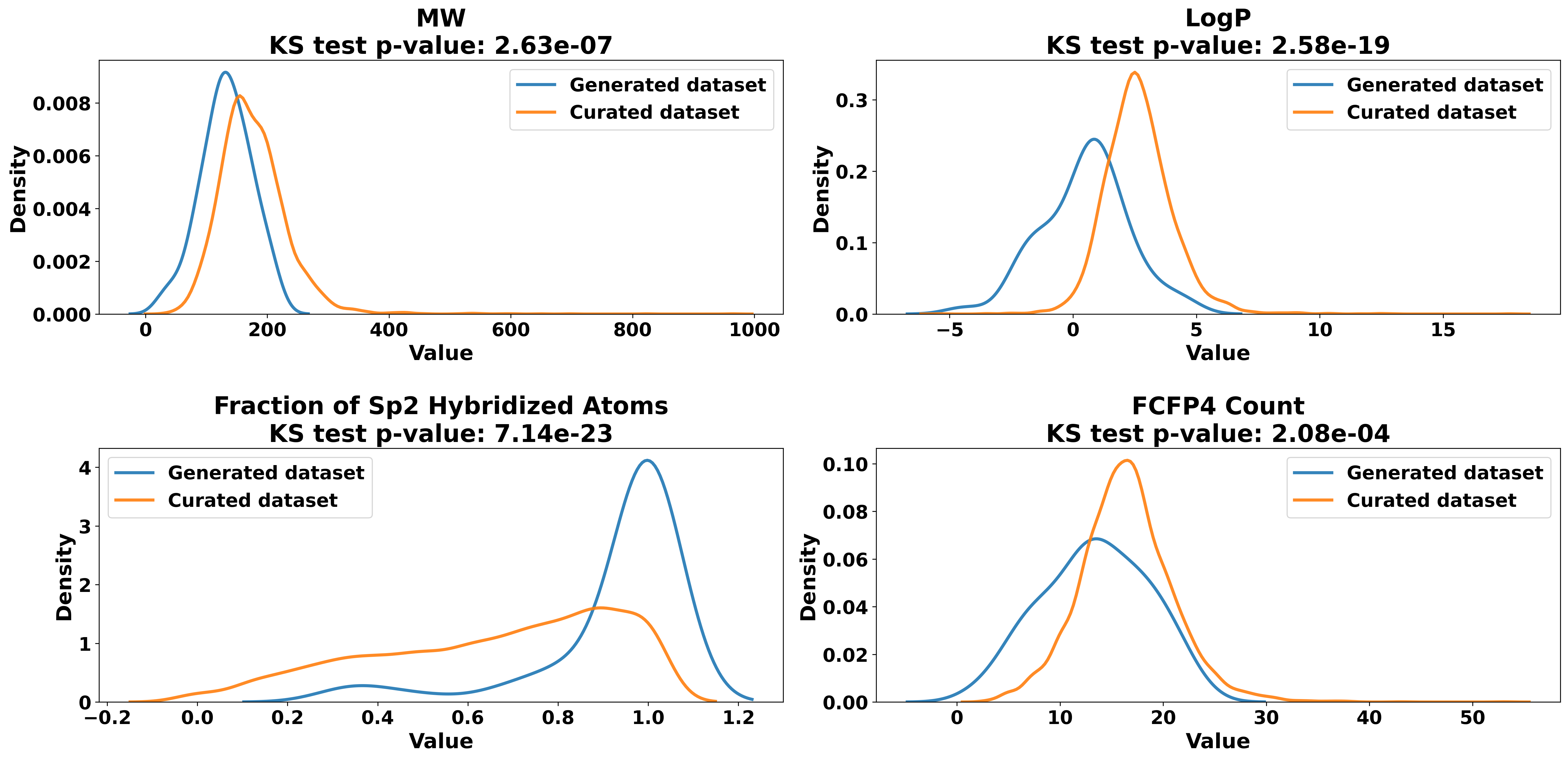}
                }
                \caption{}
                \label{fig:ks14_imagew}
            \end{subfigure}

            \vspace{1em} 

            \begin{subfigure}{\linewidth}
                \centering
                \fbox{ 
                     \includegraphics[scale=0.27]{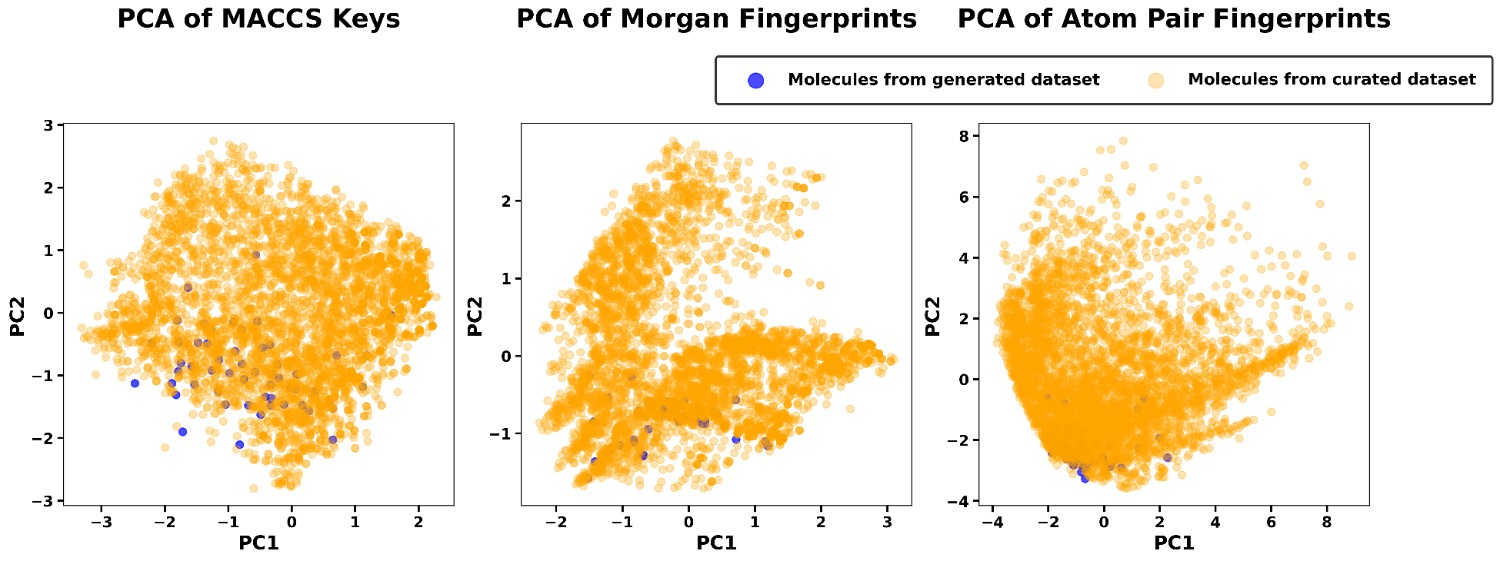}
                }
                \caption{}
                \label{fig:combinedg_fingerprints}
            \end{subfigure}
            \vspace{0.5em} 

            \begin{subfigure}{\linewidth}
                \centering
                \fbox{ 
                     \includegraphics[scale=0.20]{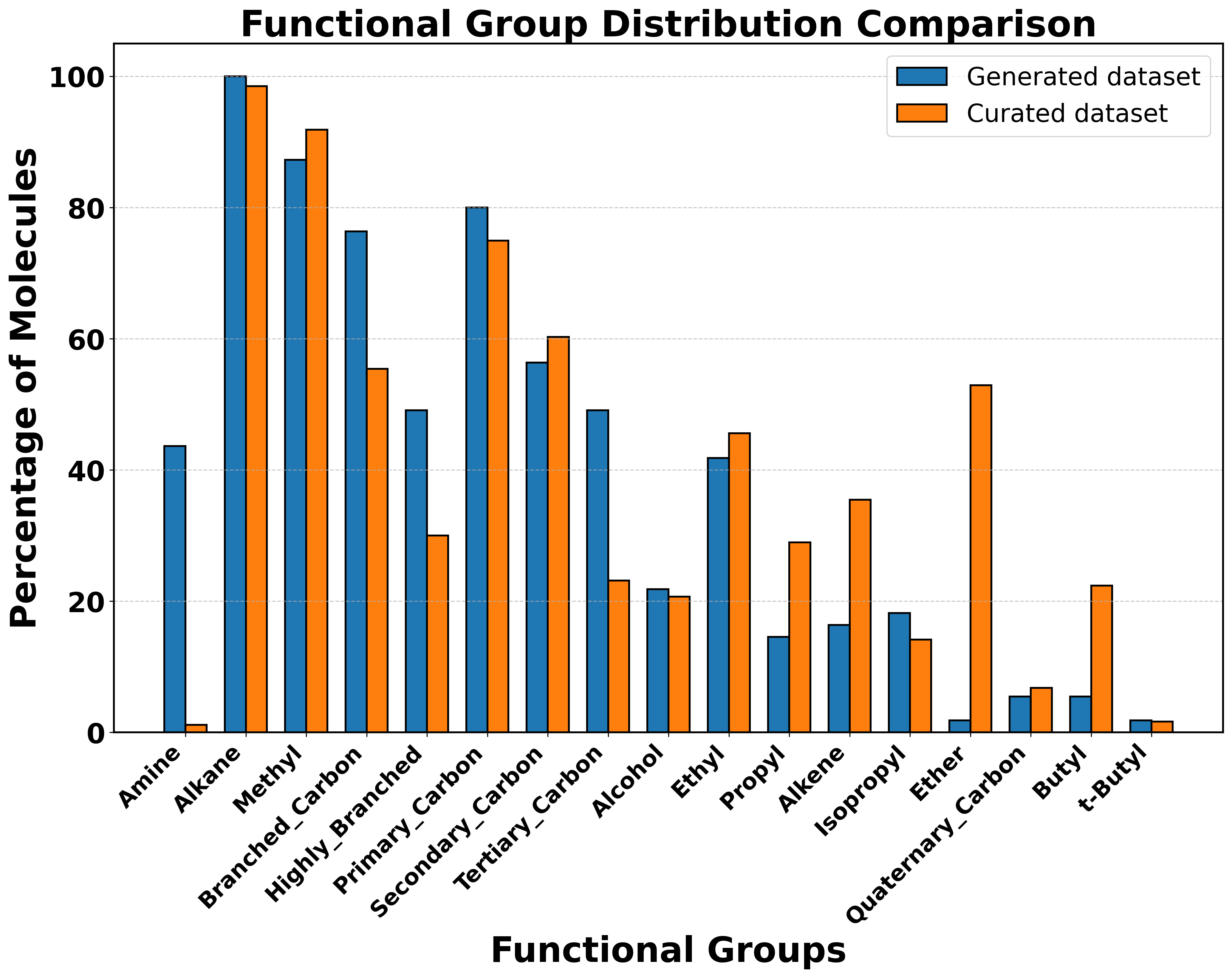}
                }
                \caption{}
                \label{fig:functional_groupsvg}
            \end{subfigure}
        \end{minipage}
    }
    
     \caption{\small\textbf{Molecular properties comparison of generated molecules from VGAE.} (a) KS test of parameters used for odor likeliness. (b) Analysis of the fingerprints. (c) Functional group analysis of the generated set.}
    \label{fig:main_figure1}
\end{figure}
\begin{figure}[htbp]
    \centering
    \fbox{%
        \includegraphics[width=1\textwidth]{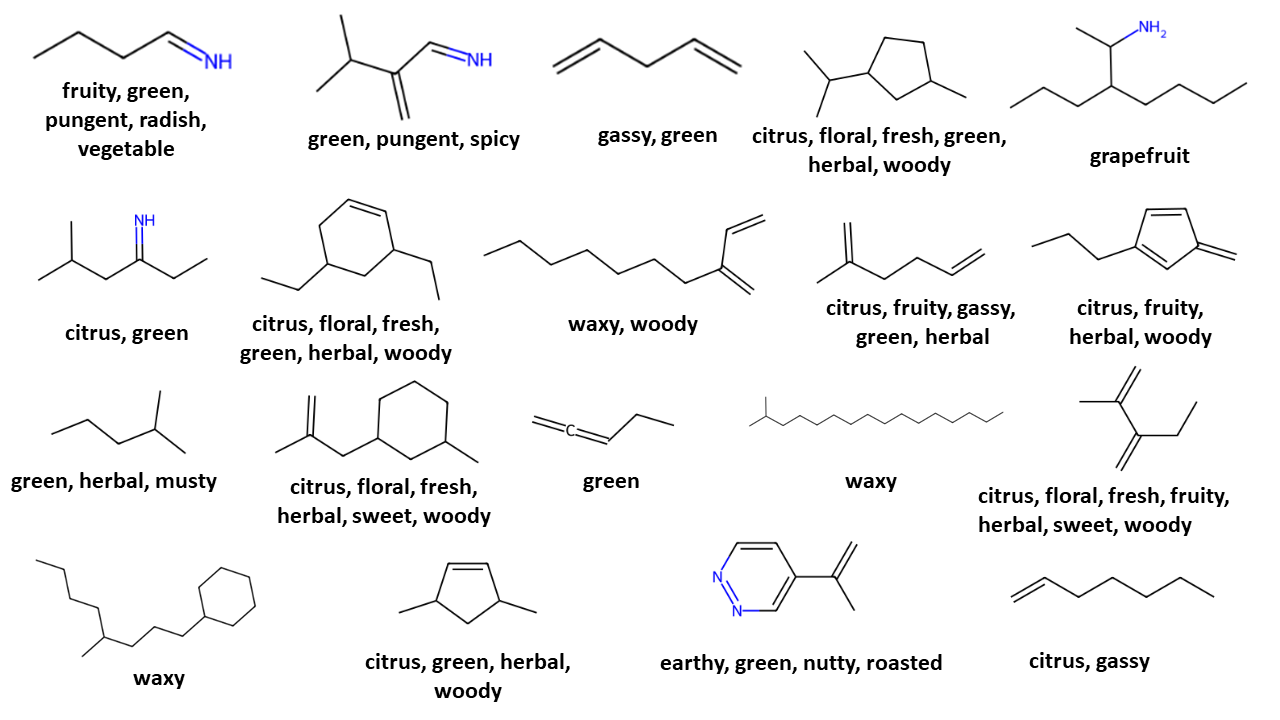}
    }
    \caption{\small\textbf{Sample of novel generated molecules(which were not in the training set) from VGAE model:} Visualization of molecular structures and their predicted odors, assigned from 138 odor labels using graph neural networks.}
    
    \label{fig:figure12}
\end{figure}
\begin{figure}[htbp]
    \centering
    \fbox{ 
        \begin{minipage}[b]{\linewidth}
            \centering
            \begin{subfigure}{\linewidth}
                \centering
                \fbox{ 
                    \includegraphics[scale=0.27]{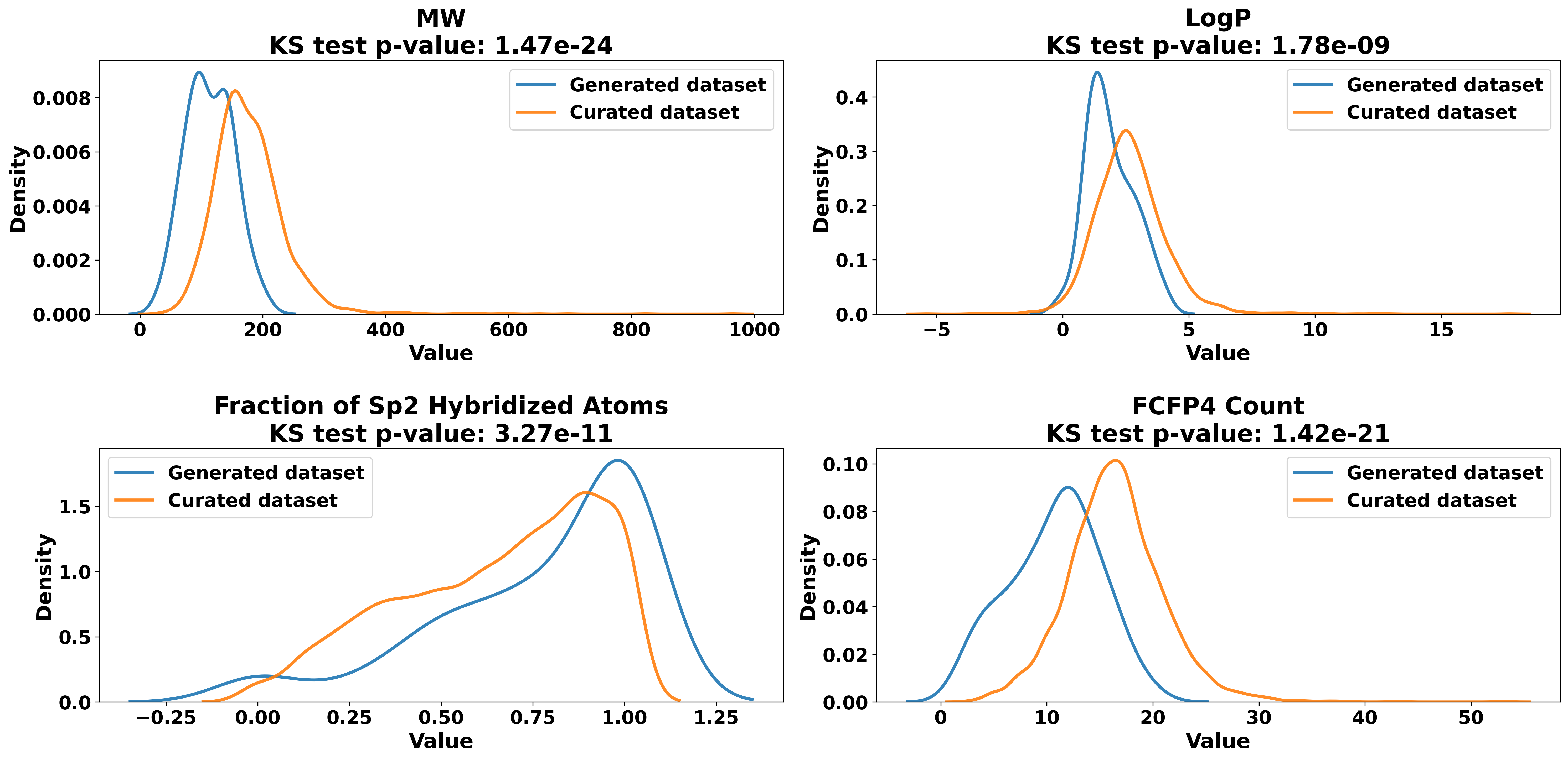}
                }
                \caption{}
                \label{fig:ks15_imagew}
            \end{subfigure}

            \vspace{1em} 

            \begin{subfigure}{\linewidth}
                \centering
                \fbox{ 
                     \includegraphics[scale=0.27]{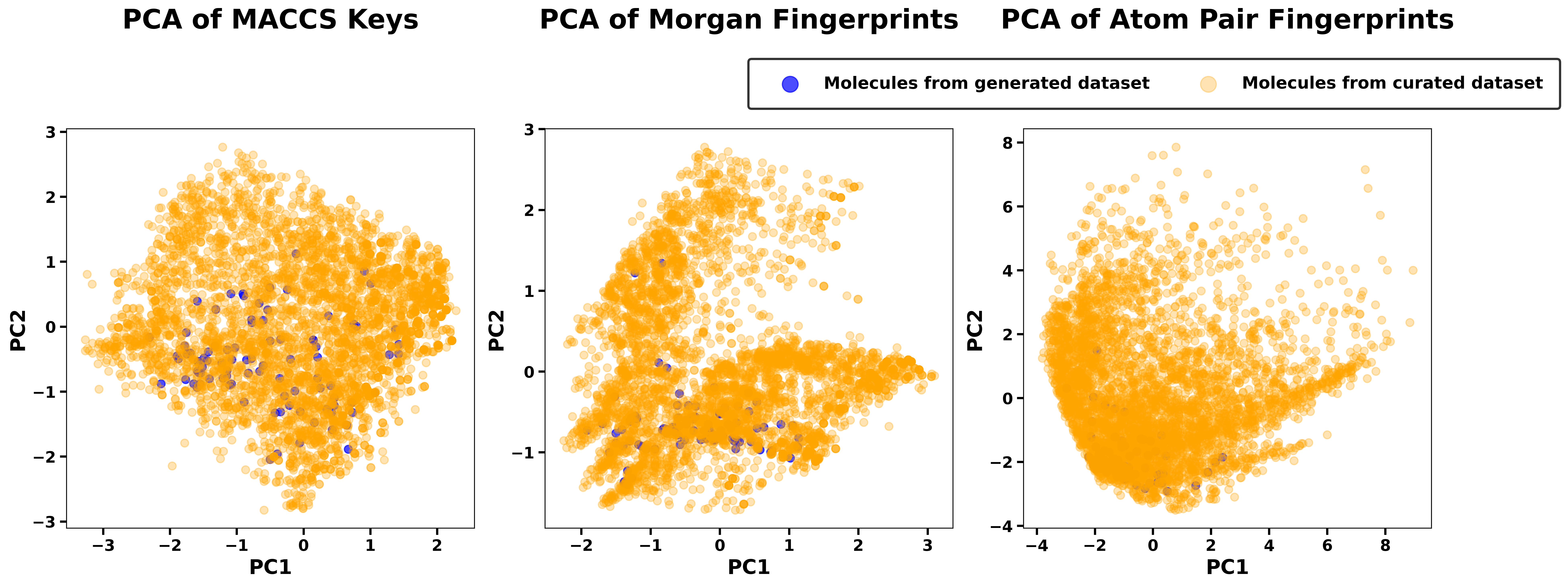}
                }
                \caption{}
                \label{fig:combineerd_fingerprints}
            \end{subfigure}
            \vspace{0.5em} 

            \begin{subfigure}{\linewidth}
                \centering
                \fbox{ 
                     \includegraphics[scale=0.20]{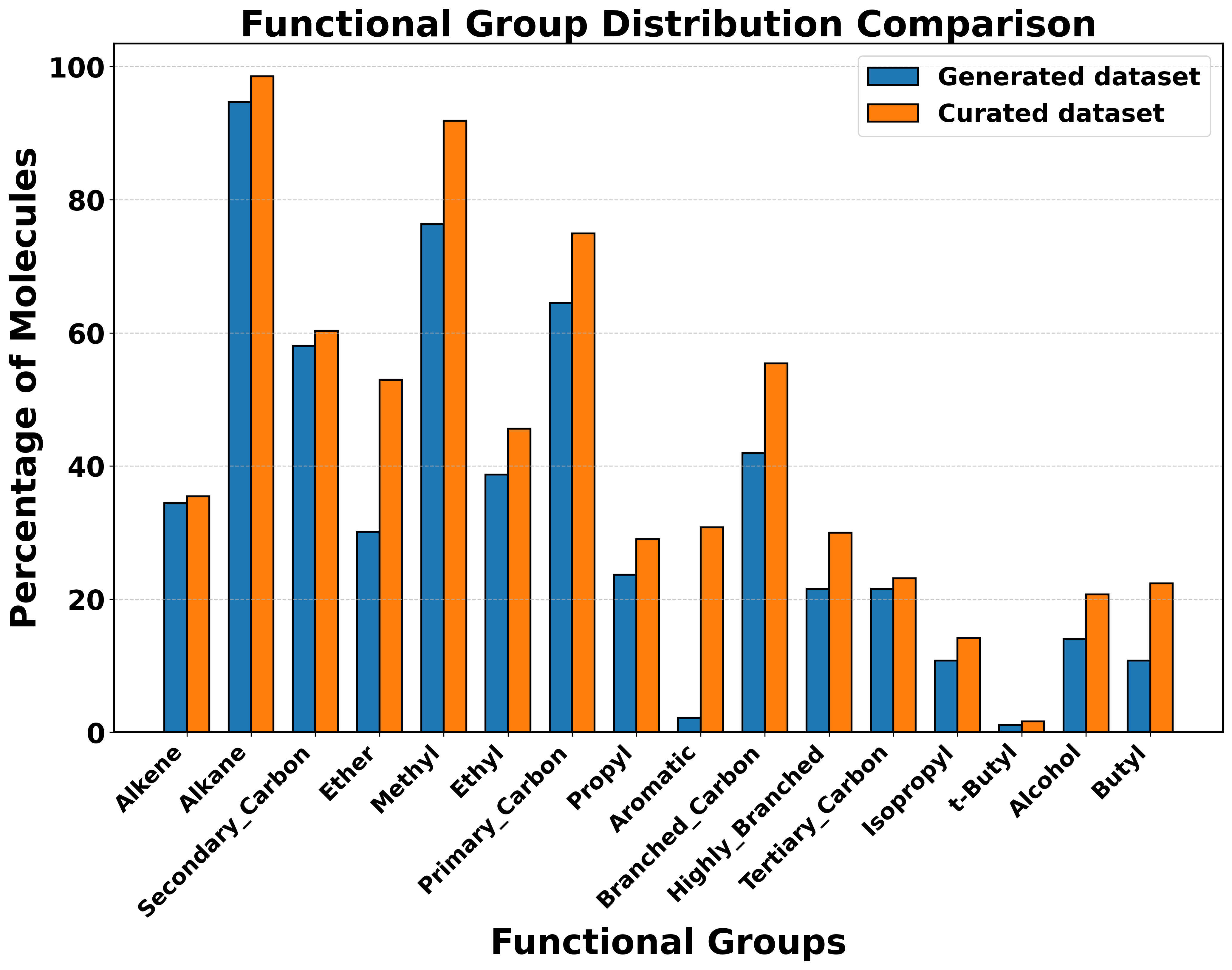}
                }
                \caption{}
                \label{fig:functional_groups1}
            \end{subfigure}
        \end{minipage}
    }
    
    \caption{\small\textbf{Molecular properties comparison of generated molecules from ARGA.} (a) KS test of parameters used for odor likeliness. (b) Analysis of the fingerprints. (c) Functional group analysis of the generated set.}
    \label{fig:main_figurearg}
\end{figure}

\begin{figure}[htbp]
    \centering
    \fbox{%
        \includegraphics[width=1\textwidth]{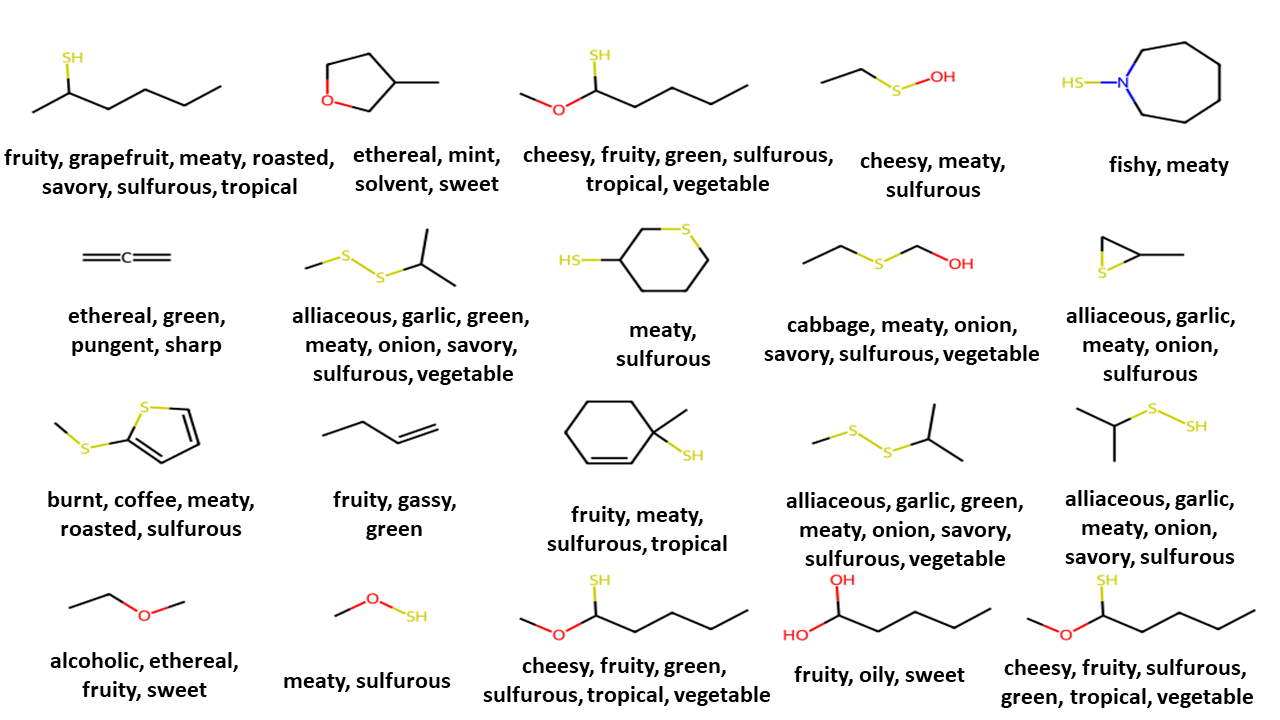}
    }
    \caption{\small\textbf{Sample of novel generated molecules(which were not in the training set) from ARGA model:} Visualization of molecular structures and their predicted odors, assigned from 138 odor labels using graph neural networks.}
    
    \label{fig:figure151}
\end{figure}

\begin{figure}[htbp]
    \centering
    \fbox{%
        \includegraphics[width=1\textwidth]{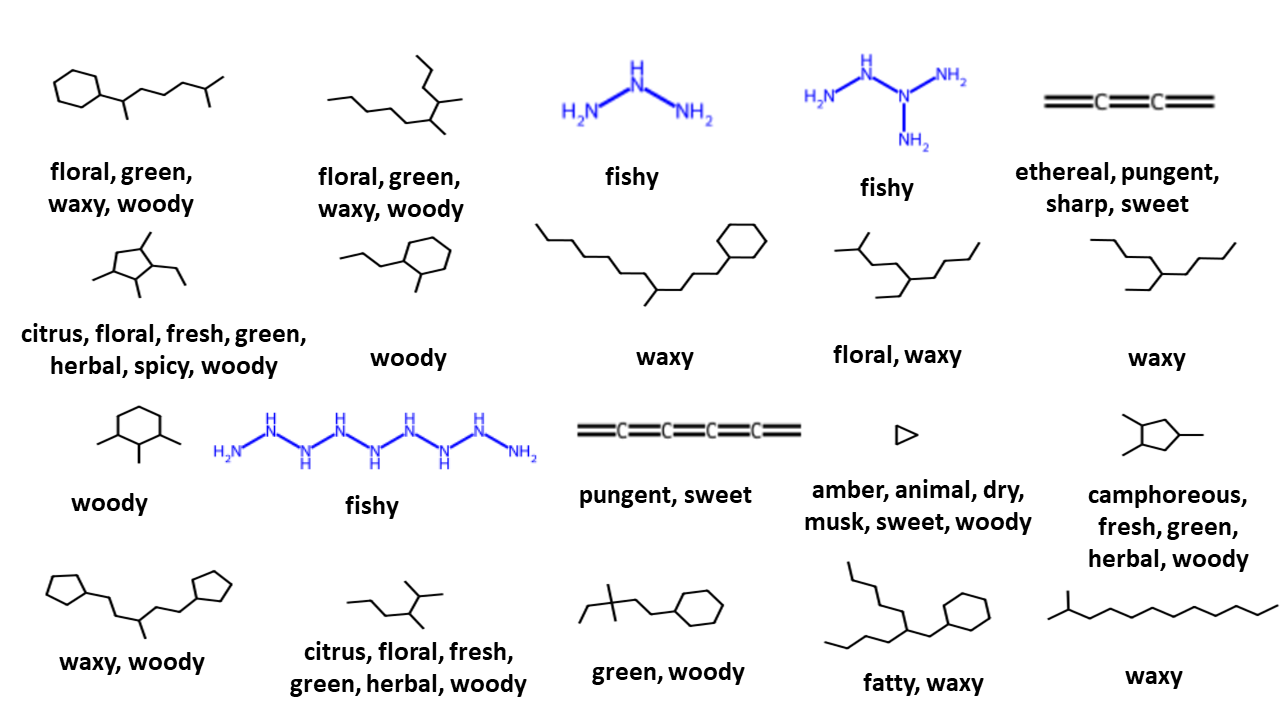}
    }
    \caption{\small\textbf{Sample of novel generated molecules(which were not in the training set) from ARGVA model:} Visualization of molecular structures and their predicted odors, assigned from 138 odor labels using graph neural networks.}
    
    \label{fig:figure171}
\end{figure}

\begin{figure}[htbp]
    \centering
    \fbox{ 
        \begin{minipage}[b]{\linewidth}
            \centering
            \begin{subfigure}{\linewidth}
                \centering
                \fbox{ 
                    \includegraphics[scale=0.27]{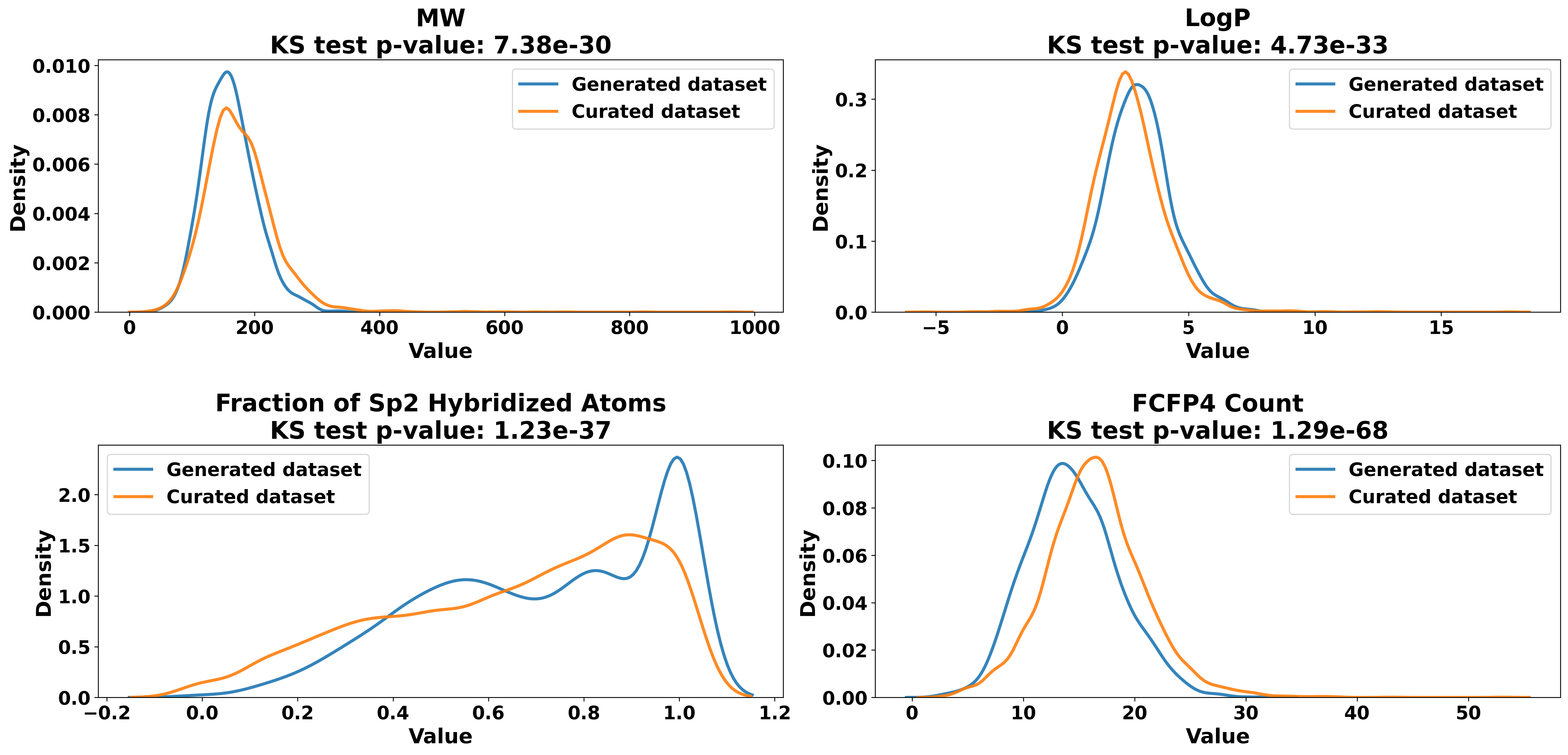}
                }
                \caption{}
                \label{fig:ks1_imagde}
            \end{subfigure}

            \vspace{1em} 

            \begin{subfigure}{\linewidth}
                \centering
                \fbox{ 
                     \includegraphics[scale=0.27]{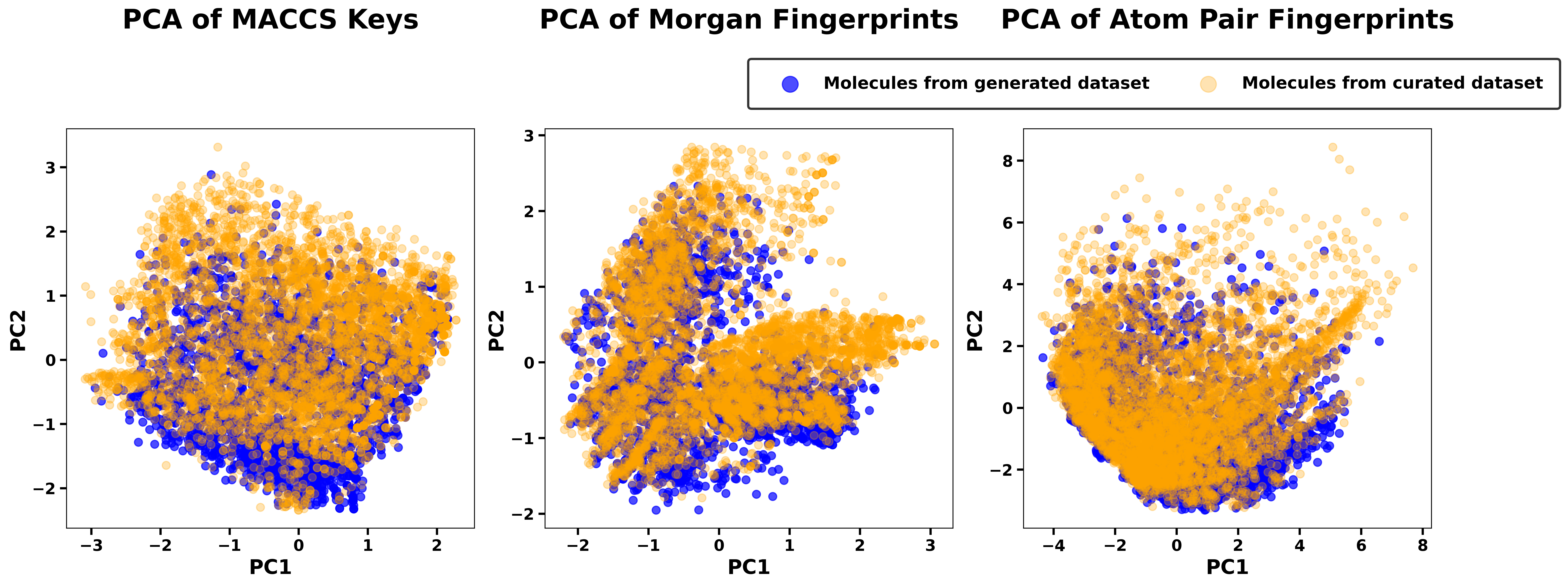}
                }
                \caption{}
                \label{fig:combretined_fingerprints}
            \end{subfigure}
            \vspace{0.5em} 

        \end{minipage}
    }
    
     \caption{\small\textbf{Molecular properties comparison of generated molecules from Diffusion.} (a) KS test of parameters used for odor likeliness. (b) Analysis of the fingerprints. (c) Functional group analysis of the generated set.}
    \label{fig:maain_figure}
\end{figure}
\begin{figure}[htbp]
    \centering
    \fbox{%
        \includegraphics[width=1\textwidth]{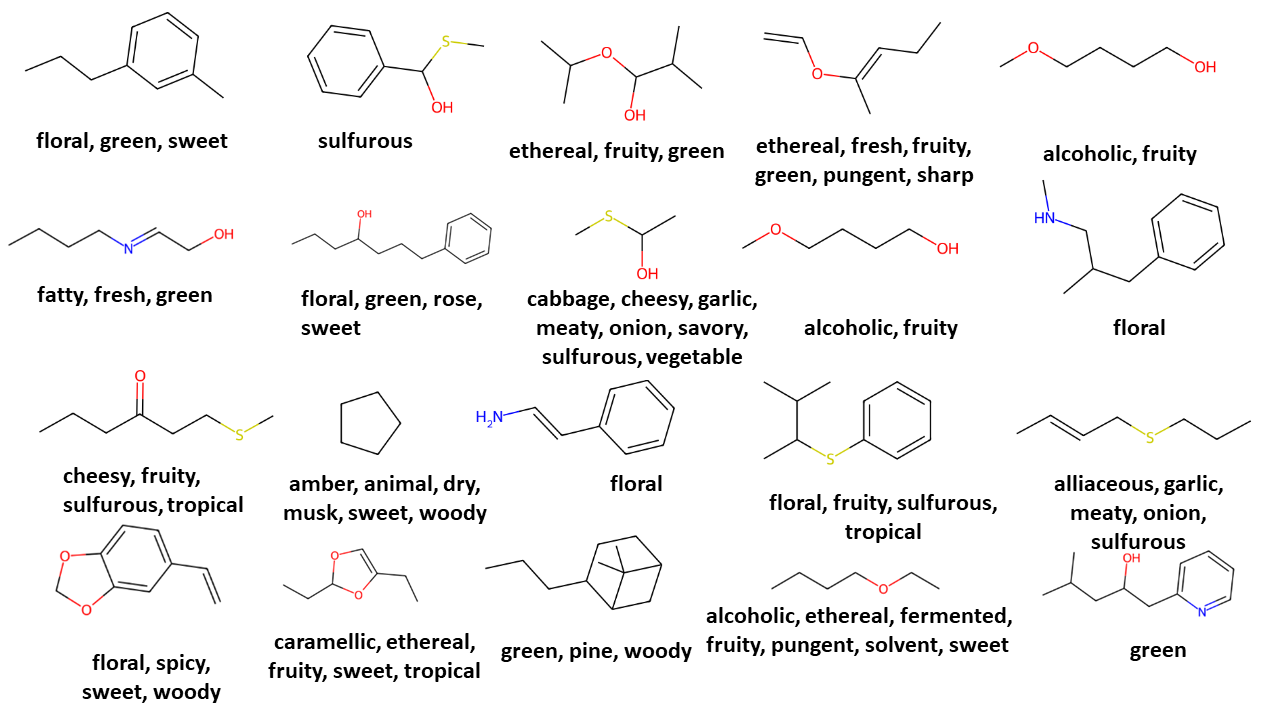}
    }
    \caption{\small\textbf{Sample of novel generated molecules(which were not in the training set) from Diffusion model:} Visualization of molecular structures and their predicted odors, assigned from 138 odor labels using graph neural networks.}
    
    \label{fig:figure11}
\end{figure}
\begin{figure}[htbp]
    \centering
    \fbox{ 
        \begin{minipage}[b]{\linewidth}
            \centering
            \begin{subfigure}{\linewidth}
                \centering
                \fbox{ 
                    \includegraphics[scale=0.27]{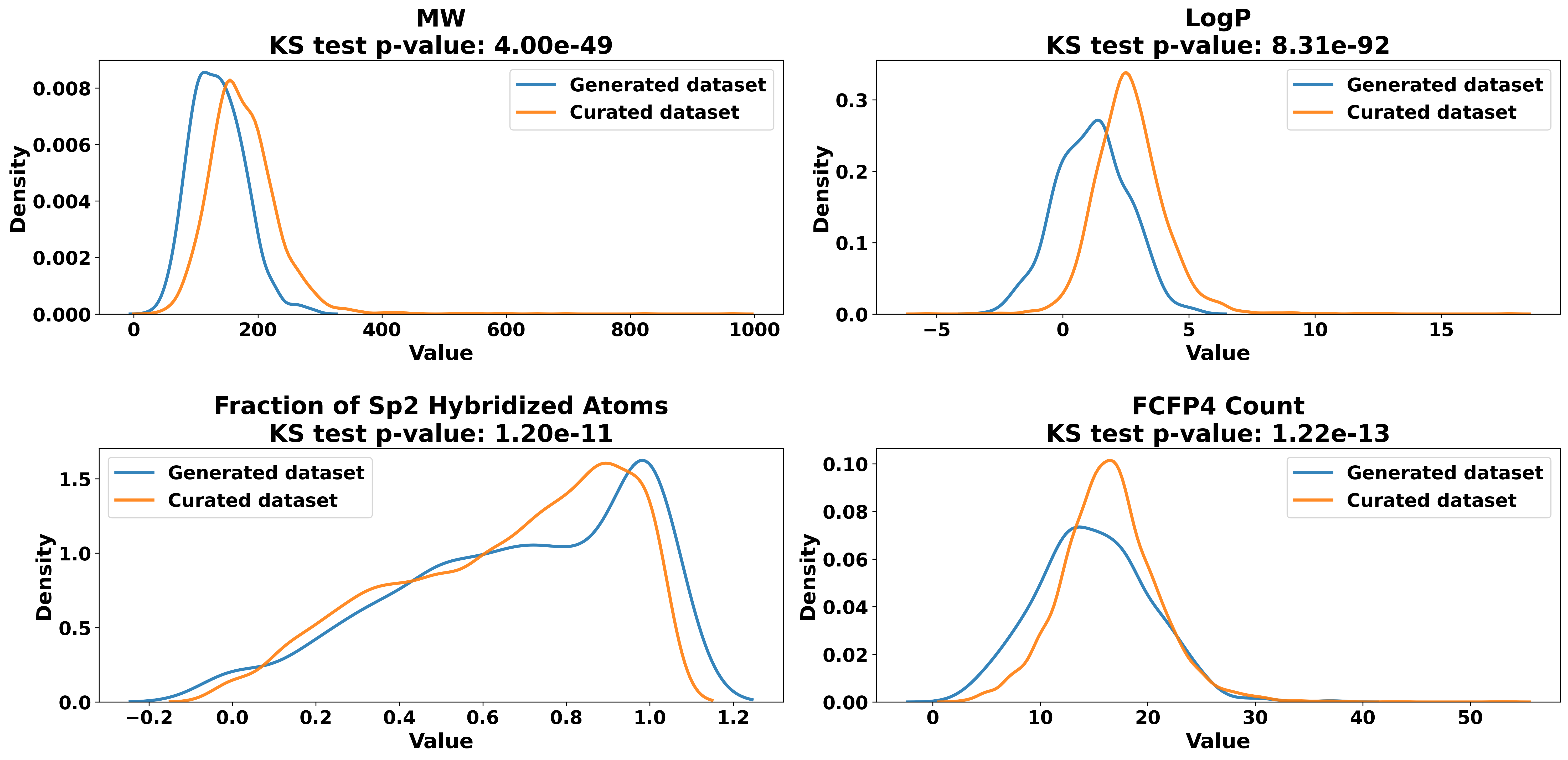}
                }
                \caption{}
                \label{fig:ks16_imaget}
            \end{subfigure}

            \vspace{1em} 

            \begin{subfigure}{\linewidth}
                \centering
                \fbox{ 
                     \includegraphics[scale=0.27]{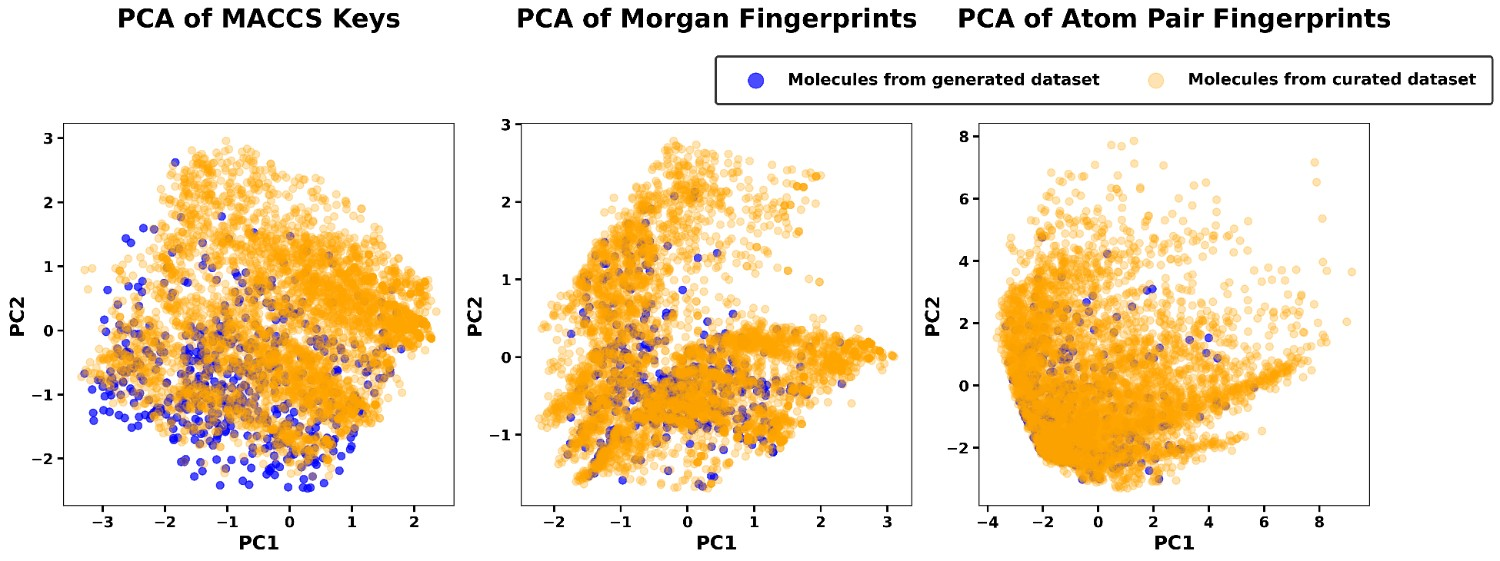}
                }
                \caption{}
                \label{fig:combined_fingerprints_t}
            \end{subfigure}
            \vspace{0.5em} 

            \begin{subfigure}{\linewidth}
                \centering
                \fbox{ 
                     \includegraphics[scale=0.20]{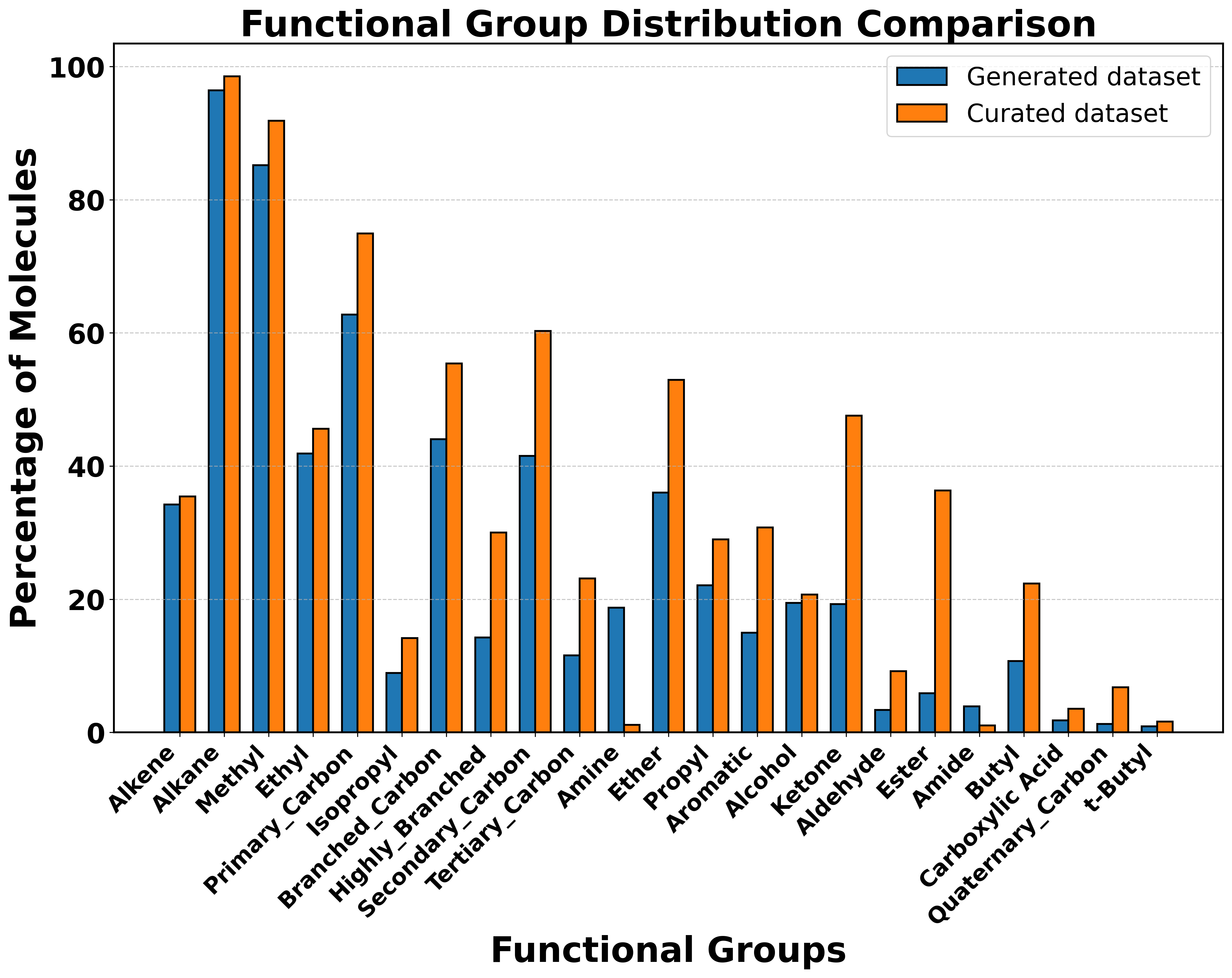}
                }
                \caption{}
                \label{fig:functionagtyl_groups}
            \end{subfigure}
        \end{minipage}
    }
    
    \caption{\small\textbf{Molecular properties comparison of generated molecules from Transformer.} (a) KS test of parameters used for odor likeliness. (b) Analysis of the fingerprints. (c) Functional group analysis of the generated set.}
    \label{fig:main_figure34}
\end{figure}


\end{document}